\documentclass[sigconf, nonacm]{acmart}

\usepackage{amsmath,amsfonts}
\usepackage{algorithmic}
\usepackage{graphicx}
\usepackage{textcomp}
\usepackage{xcolor}
\usepackage{hyperref}
\usepackage{booktabs}
\usepackage{dsfont}
\usepackage{multirow}
\usepackage{adjustbox}
\usepackage{mdframed}
\usepackage{algorithmic}
\usepackage{algorithm}
\usepackage{stfloats}
\usepackage{tikz}
\usepackage{makecell}
\usepackage{colortbl}
\usepackage{xspace}
\usepackage{enumitem}
\usepackage{subcaption}  
\usepackage{caption}

\usepackage{balance}

\newcommand{\sysname}{\texttt{CoTE-SQL}\xspace}

\newcommand\vldbdoi{XX.XX/XXX.XX}

\newcommand\vldbavailabilityurl{URL_TO_YOUR_ARTIFACTS}
\newcommand\vldbpagestyle{plain} 

\begin{document}
\title{Integrating Reasoning and Generalization in Text-to-SQL via Self-Enhanced Fine-Tuning}





\author{Feng Lyu}
\affiliation{%
  \institution{Central South University}
}
\email{fenglyu@csu.edu.cn}

\author{Jinfeng Cen}
\affiliation{%
  \institution{Central South University}
}
\email{cenjinfeng@csu.edu.cn}

\author{Sijing Duan}
\affiliation{%
  \institution{Tsinghua University}
}
\email{duansj@tsinghua.edu.cn}

\author{Hao Wu}
\affiliation{%
  \institution{Nanjing University}
}
\email{hao.wu@nju.edu.cn}

\author{Shucheng Li}
\affiliation{%
  \institution{Central South University}
}
\email{shuchengli@csu.edu.cn}

\author{Weixu Zhang}
\affiliation{%
  \institution{McGill University}
}
\email{weixu.zhang@mail.mcgill.ca}

\author{Haolun Wu}
\affiliation{%
  \institution{McGill University}
}
\email{haolun.wu@mail.mcgill.ca}

\begin{abstract}
Text-to-SQL aims to translate natural language questions into executable SQL queries over structured databases, enabling non-expert users to access data intuitively. While recent advances in large language models (LLMs) have shown promise in this task, existing LLM-based approaches often struggle to strike a balance between strong reasoning capabilities and robust generalization. To address these limitations, we propose \sysname to enhance the LLM-based text-to-SQL generation with three key innovations: (i) self-enhanced reasoning traces distilled from LLMs without human annotation, (ii) structured chain-of-thought (CoT) prompting with modular decomposition and examples retrieval, and (iii) error-aware revision based on SQL execution feedback. Extensive experiments on the Spider and Bird benchmarks demonstrate that \sysname achieves new state-of-the-art performance among methods built on open-source LLMs with comparable model sizes on Bird (53.39\% EX / 59.02 VES) and strong results on Spider (79.60\% EX / 77.19 VES), with especially significant gains on complex queries. Results highlight the effectiveness of combining self-enhancement, structured reasoning, and execution-time feedback within an LLM-based framework for text-to-SQL design. 
\end{abstract}
\maketitle
\pagestyle{\vldbpagestyle}
\ifdefempty{\vldbavailabilityurl}{}{
\vspace{.3cm}
\begingroup\small\noindent\raggedright\textbf{Artifact Availability:}\\
The source code, data, and/or other artifacts have been made available at \url{https://anonymous.4open.science/r/CoTE-SQL}.
\endgroup
}

\section{Introduction}

\begin{figure*}[t]
  \centering
  \includegraphics[width=1.95\columnwidth]{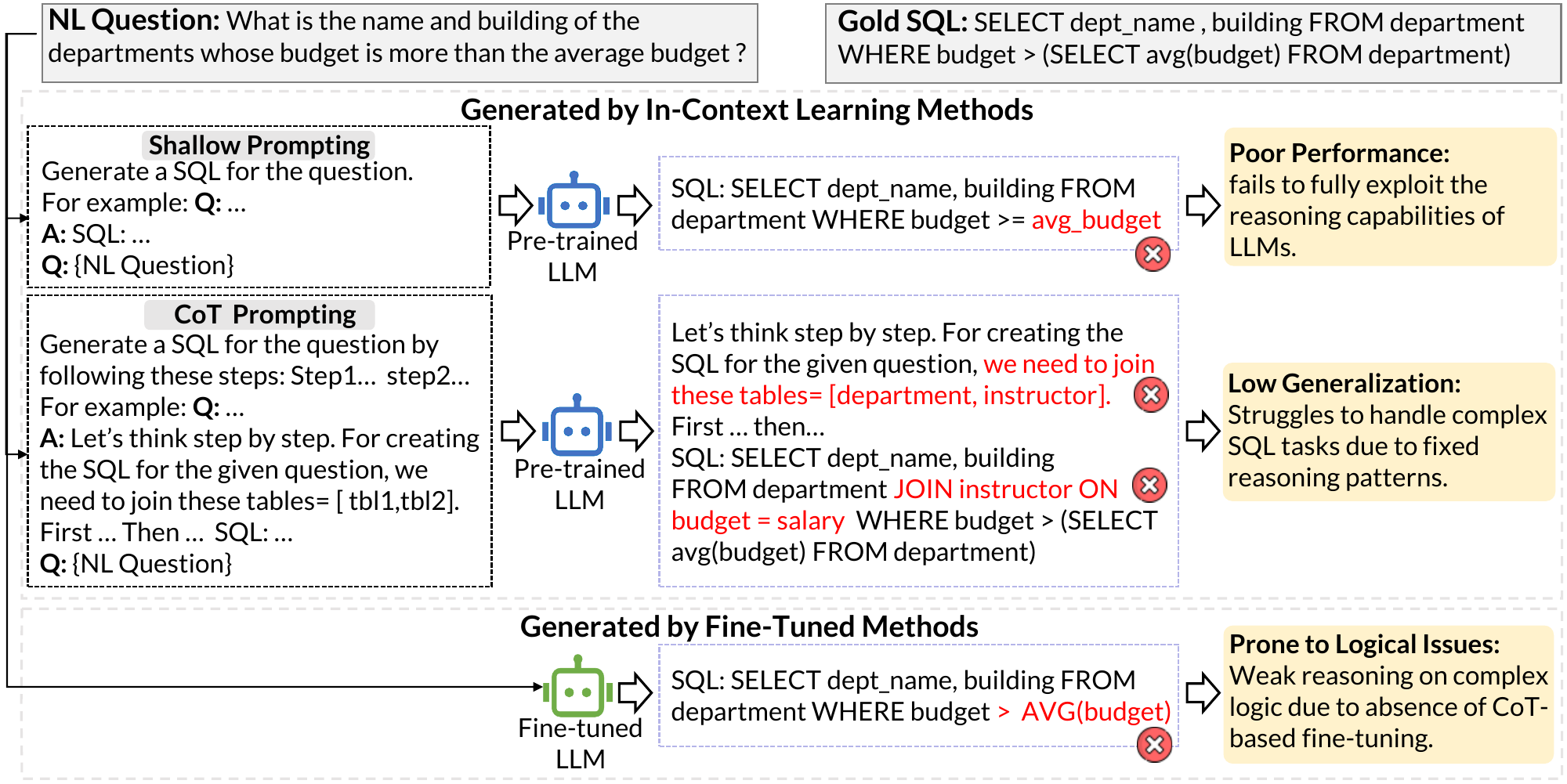}
  \caption{Limitations of existing text-to-SQL approaches. (i) In-context learning methods can be categorized into two types. The first, referred to as shallow prompting, demonstrates SQL generation via examples but fails to fully exploit the reasoning capabilities of LLMs. The second, CoT-based prompting, builds on this with CoT reasoning, but often suffers from poor generalization and struggles with complex SQL tasks due to limited reasoning diversity. (ii) Fine-tuning-based methods are prone to logical errors, constrained mainly by the quantity and quality of annotated data. Moreover, these approaches often lack explicit supervision over intermediate reasoning steps, resulting in suboptimal performance on complex queries.\label{fig:introduction}}
  \Description{}
  
\end{figure*}

As modern enterprise databases continue to expand in both scale and complexity, querying structured data remains a significant barrier for non-expert users. The traditional requirement of writing Structured Query Language (SQL) hinders broad access to relational databases, especially for users without programming expertise~\cite{shi2024survey,ren2024purple,zhao2024sphinteract}. To address this usability gap, text-to-SQL systems have emerged as a promising solution, 
focusing on the semantic conversion of information needs expressed in natural language into structured database queries
~\cite{pourreza-rafiei-2024-dts,xie-etal-2024-decomposition,NEURIPS2023_72223cc6,liu2024survey,fan2024grounding,fan2023gar,zheng2023rhb,gong2024graph}. 
This paradigm not only democratizes data retrieval but also enables new applications in data analytics and business intelligence.
Despite steady progress, accurately capturing user intent from users' natural language and generating complex, correct SQL remains a core challenge. Recent advances in large language models (LLMs) have significantly reshaped the landscape of text-to-SQL research. With their strong natural language understanding and reasoning abilities, LLMs have shown promise in bridging the gap between natural language and formal database queries~\cite{hong2024next,10530359,zhang2025clear,li2025aid,huang2024data,luo2024ptd,liu2025nl2sql}.
As reported in Figure~\ref{fig:introduction}, current LLM-based text-to-SQL approaches generally fall into two main paradigms: in-context learning and fine-tuning.

\textit{In-context learning-based methods} enhance the zero-shot or few-shot capabilities of LLMs by injecting task-specific examples into the prompt without modifying model parameters~\cite{tai-etal-2023-exploring, zhang-etal-2023-act, shen2024select}. However, these methods show poor performance because they cannot fully utilize the reasoning ability of LLMs. Therefore, some works incorporate chain-of-thought (CoT) prompting to elicit intermediate reasoning steps~\cite{xie-etal-2024-decomposition, wang2024dac, chen2025track}. However, the reliance on static, handcrafted examples or prompt templates \textit{limits the generalization ability}, especially when handling complex queries or out-of-distribution schemas.

\textit{Fine-tuning-based approaches}, in contrast, adapt LLMs to the text-to-SQL task through supervised learning on large collections of annotated (question, SQL) pairs~\cite{DBLP:conf/iclr/HuSWALWWC22, Fu_Ou_Yu_Lin_2023, hong-etal-2024-knowledge, yang-etal-2024-synthesizing}. By integrating database schema information and leveraging synthetic data augmentation, these models aim for stronger domain alignment and robustness~\cite{pourreza-rafiei-2024-dts, wang-etal-2025-mac}. However, most fine-tuning methods lack explicit supervision over intermediate reasoning processes due to the scarcity of high-quality CoT-style annotations in this domain. As a result, fine-tuned models often \textit{struggle with multi-step or compositional queries} where step-by-step logical reasoning is crucial. Although some studies fine-tune models with distilled CoT annotations~\cite{he-etal-2025-star,rossiello2025rationalization}, they often depend on strong data generators or impose rigid reasoning formats, limiting flexibility. Recent RL-based methods~\cite{pourreza2025reasoning,sheng2025csc} improve Text-to-SQL reasoning but require complex reward design and incur high training costs and instability.

To address the limitations of both in-context prompting and fine-tuning paradigms, our work seeks to bridge the gap between \textit{reasoning capability} and \textit{generalization ability} in text-to-SQL systems.


However, achieving this goal introduces several key challenges, as detailed in Section~\ref{subsec:challenges}.
\textbf{First}, the lack of explicit supervision over intermediate reasoning steps in most public datasets prevents models from learning interpretable, step-by-step derivations, which are critical for complex SQL generation. \textbf{Second}, existing models often struggle to generalize across various database schemas, due to the ambiguity of natural language and the variability of schema representations. \textbf{Third}, real-world applications impose strict correctness and reliability requirements: models must avoid spurious or hallucinated reasoning, and ideally possess the ability to self-correct without extensive manual intervention.

To tackle these challenges, we propose \sysname, a unified framework that enhances reasoning, generalization, and robustness in text-to-SQL systems with three key components (illustrated in Figure~\ref{fig:framework}).
\textit{First}, an \textbf{iterative self-enhanced fine-tuning} framework where the model generates and verifies its own reasoning traces, progressively refining its intermediate reasoning capabilities.
\textit{Second}, a \textbf{structured CoT prompting} design that guides the model through modular subtasks such as schema selection and SQL generation, augmented with retrieval-based exemplar prompts to enhance in-context learning and domain generalization.
\textit{Third}, an \textbf{error-aware revision} module that leverages execution feedback to enable self-debugging and correction of faulty SQL queries, boosting robustness in practical deployments.

We conduct extensive experiments on two standard benchmarks (Spider and Bird) to evaluate \sysname against state-of-the-art text-to-SQL methods. Results show that \sysname achieves new SOTA performance on Bird (53.39\% EX / 59.02 VES) and competitive results on Spider (79.60\% EX / 77.19 VES), outperforming 15+ baselines, including fine-tuning (e.g., MAC-SQL and DTS-SQL) and prompting methods (e.g., DAC and DIN-SQL). Notably, \sysname demonstrates consistent improvements across all difficulty levels, with particularly significant gains on the most challenging queries (7.2\% EX improvement on Spider's `Extra Hard' category).
Comprehensive ablation studies validate the contribution of each module. Human evaluation further confirms \sysname's superiority in generating complete (93/100), structurally sound (83/100), and logically consistent (86/100) SQL queries. 

In summary, we make the following key contributions:
\begin{itemize}
    \item We identify and address a gap in existing LLM-based text-to-SQL systems between reasoning capability and generalization, tackling challenges related to reasoning supervision, schema diversity, and error correction.
    \item We propose \sysname, a novel framework that integrates iterative self-enhanced fine-tuning with structured CoT prompting and error-aware revision, enabling the scalable acquisition of intermediate reasoning skills and the reliable generation of SQL.
    \item Extensive experiments on multiple challenging benchmarks demonstrate that \sysname achieves new state-of-the-art performance, significantly outperforming existing fine-tuning and prompting methods across all difficulty levels.
\end{itemize}

\section{Preliminaries}

\subsection{LLM-based Text-to-SQL}
The LLM-based text-to-SQL approach translates natural language queries into SQL statements. Given a query $Q$ and a database schema $S$ with tables $T = \{ t_1, \dots, t_{|T|} \}$, each table $t_i$ has columns $C_i = \{ c_1^{t_i}, \dots, c_{|C_i|}^{t_i} \}$.
The task is to generate an executable SQL query $Y$ that answers $Q$, modeled by estimating the conditional probability of $Y$ given the prompt $\mathcal{P} = (Q, S)$:
\begin{equation}
P_M(Y | \mathcal{P}) = \prod_{i=1}^{|Y|} P_M(Y_i \mid Y_{<i}; \mathcal{P}),
\end{equation}
where $P_M(Y_i \mid Y_{<i}; \mathcal{P})$ is the probability of generating token $Y_i$ given previous tokens $Y_{<i}$ and context $\mathcal{P}$.

\subsection{CoT Prompting for Text-to-SQL}

\textbf{CoT Prompting for Text-to-SQL.}
CoT prompting has been explored to improve LLM performance on text-to-SQL by encouraging intermediate reasoning before SQL generation~\cite{tai-etal-2023-exploring}. For a task $T$ and query $Q$, the model first derives a reasoning trace $R_{\text{task}}$:
\begin{equation}
R_{\text{task}} = \arg\max P(R \mid T, Q).
\end{equation}
The final SQL query $y$ is then produced conditioned on both the inputs and $R_{\text{task}}$.

\section{Overview}

\subsection{Problem Challenges}
\label{subsec:challenges}

Existing works on text-to-SQL fall into two extremes: either fine-tuning models without explicit reasoning supervision or using CoT-style prompting without grounding in domain-specific learning or strong generalization. To bridge this gap, we aim to develop a fine-tuning approach that explicitly enhances reasoning capabilities and improves generalization across diverse database schemas. This introduces several key challenges:


\begin{enumerate}
    \item \textbf{Lack of Reasoning-Specific Supervision}. High-quality intermediate reasoning traces are rarely available in public text-to-SQL datasets. Training solely on final SQL outputs limits the model's ability to learn structured, interpretable reasoning paths. 
    \item \textbf{Domain Generalization \& Schema Sensitivity}. LLMs often fail to generalize across various database schemas, suffering from unclear correlations and misinterpreting schema elements due to ambiguous or under-specified natural language queries.
    \item \textbf{High Correctness Requirements in Applications}. In practical deployments, LLMs must avoid generating incorrect or hallucinated reasoning paths. Effective systems must detect, revise, and learn from their errors without relying on manual supervision.
\end{enumerate}

\begin{figure}[t]
  \centering
\includegraphics[width=0.88\linewidth]{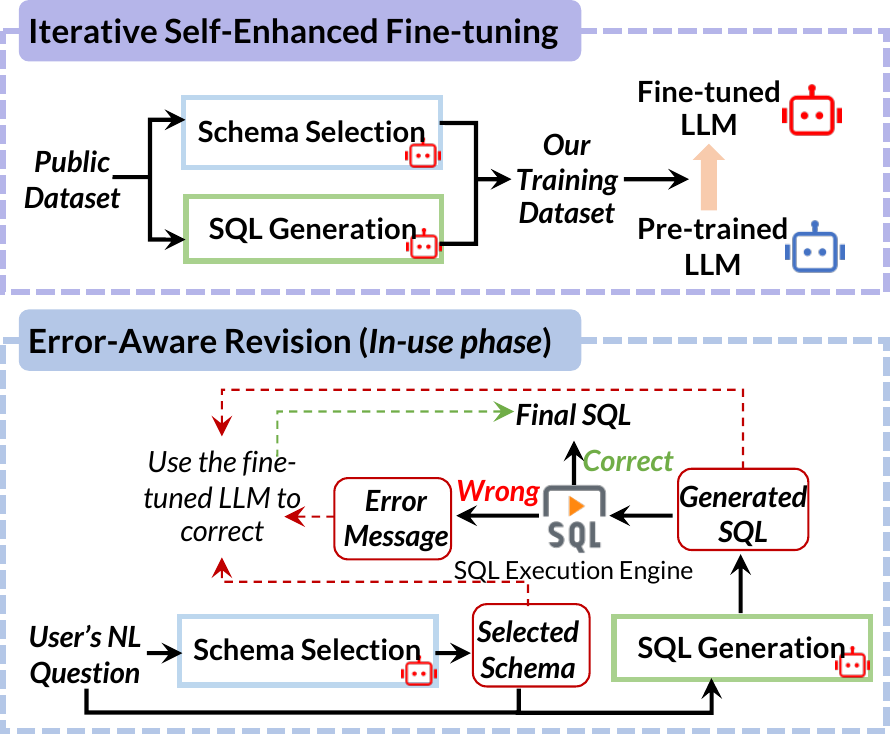}
  \caption{Overview of \sysname. The upper part illustrates the self-enhanced fine-tuning approach, while the lower part shows the error-aware revision during inference. Schema selection and SQL generation are used in both the fine-tuning and inference phases.\label{fig:framework}}
  \Description{}
\end{figure}

\subsection{Core Insights}

To address these challenges, our design is driven by three key insights:

\begin{enumerate}
    \item \textbf{Self-Enhanced Reasoning from LLMs}. Manually constructing high-quality reasoning traces is expensive and non-scalable. Inspired by recent self-training techniques, we can propose a self-enhanced fine-tuning framework that extracts latent reasoning traces from LLMs themselves. By aligning these traces with gold answers and filtering for correctness, we construct high-fidelity supervision without expert annotations.

    \item \textbf{Structured CoT Prompt for Maximizing Reasoning Ability}. We aim to design a flexible yet structured CoT prompting framework that decomposes the text-to-SQL task into modular reasoning stages, such as schema linking and SQL planning. This structured decomposition enhances interpretability while maintaining generation flexibility. Furthermore, we can augment prompts with retrieval-based exemplars from a curated set of text-SQL pairs, improving in-context learning and adaptation to unseen schemas.
    
    \item \textbf{Error-Driven Reasoning Revision}. Instead of discarding failed outputs, we can treat SQL execution errors as weak supervision signals. By pairing error messages with reflective CoT prompts, we can guide the model to analyze its own failure points and iteratively revise the SQL, a form of self-debugging driven by runtime feedback.
    
    
\end{enumerate}

\begin{figure*}[t]
  \centering
  \includegraphics[width=0.95\linewidth]{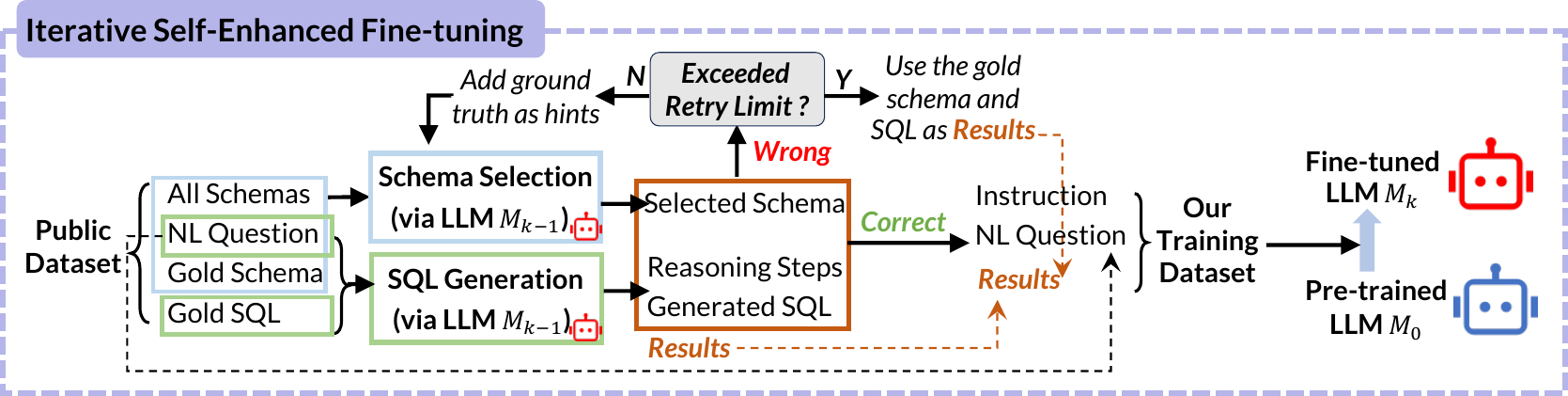}
  \caption{The proposed iterative self-enhanced fine-tuning framework. Note that in practical use, \sysname evaluates the correctness of the selected schema and the generated SQL separately. If one is incorrect, only the corresponding module needs to be retried. We plot them together for simplicity.\label{fig:framework_self_enhance}}
  \Description{}
\end{figure*}

\subsection{Key Designs}

Based on the insights above, we propose \sysname, whose workflow is illustrated in Figure~\ref{fig:framework}. It comprises the following key design components:

\textbf{1. Iterative Self-Enhanced Fine-tuning Framework (\S\ref{subsec:self_enhanced_ft}).}
We propose a multi-round, self-enhanced training framework where the LLM acts not only as a reasoning engine but also as a generator of its own intermediate reasoning traces. At each iteration, the model is prompted to produce reasoning steps that align with the gold SQL answers. These outputs are filtered based on execution correctness or logical alignment, and only verified traces are retained as training data. Incorrect outputs are augmented with hints and reprocessed to improve coverage. Notably, the reasoning traces used in this framework are generated using our structured CoT design (\S\ref{subsec:cot_design}), ensuring structure and interpretability. This iterative self-supervision pipeline enables the model to gradually acquire robust, domain-adapted reasoning skills, without relying on external labels or teacher models.

\textbf{2. Structured CoT with Modular Reasoning and Retrieval Cues (\S\ref{subsec:cot_design}).}
To reliably generate training-quality reasoning traces, we design a structured CoT prompting strategy that divides the text-to-SQL task into modular subtasks. Each module, such as \textit{schema selection} or \textit{SQL generation}, uses role-specific instructions, localized guidance, and annotated demonstrations. In contrast to rigid template-based methods, our design supports free-form, self-explanatory reasoning within a scaffolded structure. To further enhance adaptability, we incorporate retrieval-augmented in-context learning by selecting soft exemplar pairs based on question similarity. These structured, retrieval-aware CoT traces not only enhance generation accuracy but also provide the foundation for self-improving fine-tuning.

\textbf{3. Error-Aware Revision with Execution-Guided Debugging (\S\ref{subsec:error_aware_revision}).}
To meet real-world accuracy standards, we introduce a SQL correction module that transforms runtime execution errors into learning signals. When a generated SQL query fails to execute, the system pairs the returned error message with a specialized CoT prompt that instructs the LLM to analyze the failure, identify the error source, and revise the SQL accordingly. This process mimics self-debugging: the model reasons through the error, revises both syntax and logic, and re-validates the output. Common error types, such as invalid operators, ambiguous fields, or syntax violations, are handled through detailed, stage-wise correction instructions. These corrected outputs not only enhance robustness during inference but are also recycled into the training loop to reinforce reasoning over time.

Note that the structured CoT prompt serves as the base for both fine-tuning and inference. Specifically, it provides the prompting design used in the self-enhanced fine-tuning phase to generate interpretable reasoning traces with selected schema and generated SQL. Also, it guides the error-aware revision process during inference. By unifying prompting strategies across both phases, this design ensures consistency, modularity, and transferability of reasoning behaviors.

\section{System Design}

\subsection{Iterative Self-Enhanced Fine-tuning Framework}
\label{subsec:self_enhanced_ft}

Pre-trained LLMs lack the domain-specific reasoning required for accurate text-to-SQL generation in \sysname. Although fine-tuning offers performance gains, it is fundamentally limited by the absence of intermediate reasoning traces as supervision. To fully unlock the generalizable reasoning potential of LLMs, we introduce a framework that enables the model to iteratively generate and validate its own reasoning traces as high-quality supervision.

As illustrated in Figure~\ref{fig:framework_self_enhance}, the proposed framework performs multiple rounds of self-enhanced fine-tuning. At each iteration, the LLM produces CoT reasoning aligned with gold SQL answers, guided by a structured CoT prompt (\S\ref{subsec:cot_design}). Generated traces are filtered using task-specific correctness criteria: schema selection outputs are validated via exact entity alignment matching in JSON format, while SQL generation outputs are verified through execution equivalence with the gold queries.

Formally, let $D^{gen} = \{(x_i^{gen}, y_i)\}_{i=1}^{N}$ denote the generation dataset, where $x_i^{gen}$ is the CoT prompt input and $y_i$ the corresponding gold answer. At iteration $k$, the current model $M_{k-1}$ produces output $o_i$ from $x_i^{gen}$, from which we extract the reasoning trace $r_i$ and predicted answer $\hat{y}_i$. If $\hat{y}_i$ satisfies the correctness indicator $\mathds{1}(y_i, \hat{y}_i) = 1$, the pair $(x_i^{gen}, o_i)$ is incorporated into the fine-tuning dataset $D^{train}$. Otherwise, the input is augmented with a corrective hint $h$ to form $x_i^{gen_h}$, and the model is prompted again to generate $o_i$. This retry mechanism continues up to a maximum number of attempts to improve coverage and robustness.

To mitigate distribution shift during training and inference, pairs involving hinted inputs $(x_i^{gen_h}, o_i)$ are sampled with a smaller probability (e.g., 0.1), as such hints are unavailable at test time. For samples that fail to generate correct reasoning after all retries, we add $(x_i^{gen_h}, y_i)$ to $D^{train}$, ensuring the model learns from difficult cases and enhancing data diversity.

Unlike conventional self-training approaches that fine-tune on the model from the previous iteration, we fine-tune the original pre-trained LLM $M_0$ with the accumulated $D^{train}$ at each iteration, thus preventing error amplification. The objective optimized during fine-tuning is the standard autoregressive token-level cross-entropy loss:

\begin{equation}
\mathcal{L} = \mathbb{E}_{(x,y) \sim D^{train}} \left[-\sum_{t=1}^T \log P(y_t | y_{<t}, x) \right],
\end{equation}
where $T$ denotes the sequence length, and $(x,y)$ a training pair.

Iterating this process for $K$ rounds yields progressively stronger models $\{M_k\}_{k=1}^K$ with enhanced domain-adapted reasoning capabilities, benefiting from higher-quality datasets in each round. This iterative self-enhancement strategy not only alleviates the reliance on external supervision but also enables effective adaptation to new database schemas or domains, as detailed in Algorithm~\ref{alg:self_enhance}.


\begin{algorithm}[ht]
\caption{Iterative Self-Enhanced Fine-tuning\label{alg:self_enhance}}

\renewcommand{\algorithmicrequire}{\textbf{Input:}}
\renewcommand{\algorithmicensure}{\textbf{Output:}}
\begin{algorithmic}[1]
\REQUIRE Pre-trained LLM $M_0$, generation dataset $D^{gen}=\{(x_i^{gen}, y_i)\}_{i=1}^N$, number of iterations $K$.
\ENSURE Fine-tuned LLM $M_K$.
\STATE Initialize $M \leftarrow M_0$, $D^{train} \leftarrow \emptyset$
\FOR {$k=1$ to $K$}
    \FOR {each $(x_i^{gen}, y_i)$ in $D^{gen}$}
        \STATE Generate $o_i = M(x_i^{gen})$, extract $(r_i, \hat{y}_i)$
        \STATE retries $\leftarrow 0$
        \WHILE{retries $< max\_retries$ and $\mathds{1}(y_i, \hat{y}_i) = 0$}
            \STATE Augment $x_i^{gen}$ with hint $h$ to get $x_i^{gen_h}$
            \STATE Generate $o_i = M(x_i^{gen_h})$, extract $(r_i, \hat{y}_i)$
            \STATE retries $\leftarrow$ retries + 1
        \ENDWHILE
        \IF{$\mathds{1}(y_i, \hat{y}_i) = 1$}
            \STATE Add $(x_i^{gen}, o_i)$ to $D^{train}$ with probability $0.9$
            \STATE Add $(x_i^{gen_h}, o_i)$ to $D^{train}$ with probability $0.1$
        \ELSE
            \STATE Add $(x_i^{gen_h}, y_i)$ to $D^{train}$
        \ENDIF
    \ENDFOR
    \STATE Fine-tune $M_0$ on $D^{train}$ to obtain $M_k$
\ENDFOR
\RETURN $M_K$
\end{algorithmic}
\end{algorithm}

Our iterative self-enhanced fine-tuning framework systematically extracts and refines high-fidelity intermediate reasoning traces from the model itself, thereby overcoming the scarcity of explicit supervision and significantly boosting the interpretability and accuracy of \sysname's text-to-SQL reasoning.

Given the inherent complexity of SQL generation, including handling nested queries, joins, aggregations, and conditionals, we allocate a larger proportion of training data to the SQL generation module compared to schema selection. Schema selection primarily involves identifying relevant tables and columns, which is a comparatively straightforward process. This balanced data allocation strategy ensures both modules achieve robust performance.

\subsection{Structured CoT Prompt Design}
\label{subsec:cot_design}

To address the challenge of domain generalization and schema sensitivity in text-to-SQL tasks, we propose a structured CoT prompt that integrates modular decomposition with retrieval-based in-context examples. 
Different from rigid template-based prompting, our design introduces a flexible reasoning interface that preserves the LLM's generative freedom while enforcing structure. 
By segmenting the reasoning process into interpretable subtasks, including schema selection (\S\ref{subsubsec:schema_selection}) and SQL generation (\S\ref{subsubsec:sql_generation}), and augmenting prompts with examples tailored to both the schema and the query, our method ensures robust adaptation to unseen databases and natural language variations. This design not only improves inference-time generation but also yields fine-tuning quality reasoning traces for a self-improving training loop.


\subsubsection{CoT Prompt Template}
\label{subsubsec:prompt_template}

At the heart of our framework is a unified prompt design that enables stepwise reasoning across diverse modules.\footnote{The full prompt templates can be found in our source code.} Each prompt follows a consistent template comprising:
\textbf{i)~Role specification} to contextualize the LLM's function (e.g., schema linker or SQL planner).
\textbf{ii)~Reasoning instructions} to clarify the expected format and process.
\textbf{iii)~Few-shot examples} drawn from curated demonstrations aligned with question similarity. 
\textbf{iv)~Problem input} appended with the cue phrase \textit{``Let's think step by step''} to activate CoT reasoning.


In contrast to prior work that constrains reasoning within rigid formats~\cite{xie-etal-2024-decomposition, NEURIPS2023_72223cc6, zhang-etal-2023-act}, we allow free-form yet guided reasoning by carefully decoupling demonstration format from execution logic. This approach empowers LLMs to internalize task structures while expressing reasoning naturally.


\subsubsection{Schema Selection via Entity Alignment}
\label{subsubsec:schema_selection}

Schema selection is crucial in mitigating the schema sensitivity of LLMs. Given a natural language query, only a subset of the database schema is typically relevant. Including irrelevant tables or columns can dilute the LLM's attention and lead to erroneous reasoning in downstream modules.

\begin{figure}[t]
  \centering
    \includegraphics[width=1\linewidth]{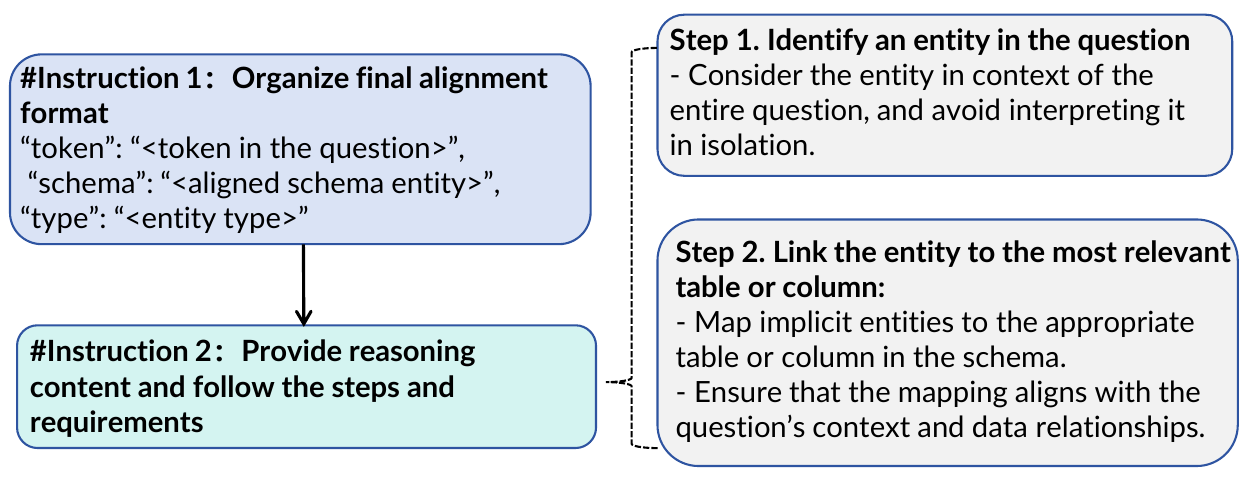}
      \caption{Reasoning instruction for schema selection.\label{fig:instru_schema_select}}
      
      \Description{}
\end{figure}

To address this, we design a schema grounding module that explicitly links natural language entities to database elements. As illustrated in Figure~\ref{fig:instru_schema_select}, this is operationalized as a CoT task with two substeps:
\textbf{(i)~Entity Identification}:  extracting tokens or phrases referencing data concepts in the query, and
\textbf{(ii)~Entity Linking}: aligning them with relevant tables and columns in the schema.

Unlike surface-level linking, our prompt explicitly models the compositionality and ambiguity of entity references. For instance, in the query  \textit{``Show the name and release year of the song by the youngest singer,''} a naïve LLM may incorrectly associate \textit{``name''} with the singer rather than the song. Our prompt includes detailed annotations and localized guidance (e.g., reminders about referential cues such as \textit{``who''}, \textit{``when''}, etc.) to reduce such errors. An example of the schema selection process is shown in Figure~\ref{fig:case_schema_select}.


\begin{figure}[t]
  \begin{tikzpicture}
  \node (table) [inner sep=0pt] {
  \begin{adjustbox}{width=0.9\linewidth}
    \begin{tabular}{l} 
     \\ [-2.0ex]
     \rowcolor{gray!10}\makecell[l]{
     \textbf{Input:} \\
     \emph{Schemas}: \textcolor{purple}{CREATE TABLE} ``singer" (``Singer\_ID" int, \\
     ``Name" text, ``Country" text, ``Song\_Name" text, \\ ``Song\_release\_year" text, ``Age" int, ``Is\_male" bool, \\
     \textcolor{purple}{PRIMARY KEY} (``Singer\_ID");) \dots\\
     \emph{Question}: \textit{How many singers are from each country? }\\
     \\[-1.0ex]
     \textbf{LLM Output:} \\
    \textit{To solve this problem, let's break it down step by step.}\\
    \textit{1. Identify the entity in the question: }\\
    \textit{The entity in the question is ``singers" and ``country"}\\
    \textit{2. Link the entity to the most relevant table or column: }\\
    \textit{The entity ``singers" is related to the table ``singer". }\\ 
    \textit{The entity ``country" is related to the column }\\
    \textit{``Country" in the table ``singer" }\\
    \textit{\dots }\\
    \textit{Based on the above analysis, the correct alignments are: }\\
    \textit{\text{[}\{``token": ``singers", ``schema": ``singer", ``type": ``tbl"\}, }\\
    \textit{\{"token": ``country", ``schema": ``singer.Country", ``type": ``col"\}\text{]} }\\  
     }
  \end{tabular}
  \end{adjustbox}
  };
  \end{tikzpicture}
   \caption{Demonstration of the schema selection.\label{fig:case_schema_select}}
   \Description{}
\end{figure}

Each alignment output is structured as a list of mappings, including the original token, the corresponding schema element, and the entity type (table or column). These structured outputs not only guide schema pruning but also serve as pseudo-labels for fine-tuning. As a result, we reduce schema noise while enhancing generalization to unseen databases.


\subsubsection{Progressive SQL Generation with CoT Planning}
\label{subsubsec:sql_generation}

The final module in our pipeline is responsible for SQL generation. Rather than generating queries in a single step, we adopt a progressive CoT strategy that mimics human-like planning. The prompt guides the LLM through stages such as identifying the intent, selecting relevant operations (e.g., \texttt{SELECT}, \texttt{GROUP BY}), and determining compositional constructs (e.g., joins, subqueries).


This progressive decoding framework supports adaptive reasoning:
For \textbf{simple queries}, the LLM may bypass redundant steps after evaluating task simplicity. 
For \textbf{complex queries}, the LLM incrementally decomposes the task, preserving contextual consistency without requiring additional prompt orchestration.
Note that the simplicity evaluation is implicit, as it relies on the model itself to determine, during the reasoning process, whether further steps are necessary. 



\begin{figure}[t]
  \centering
\includegraphics[width=1\linewidth]{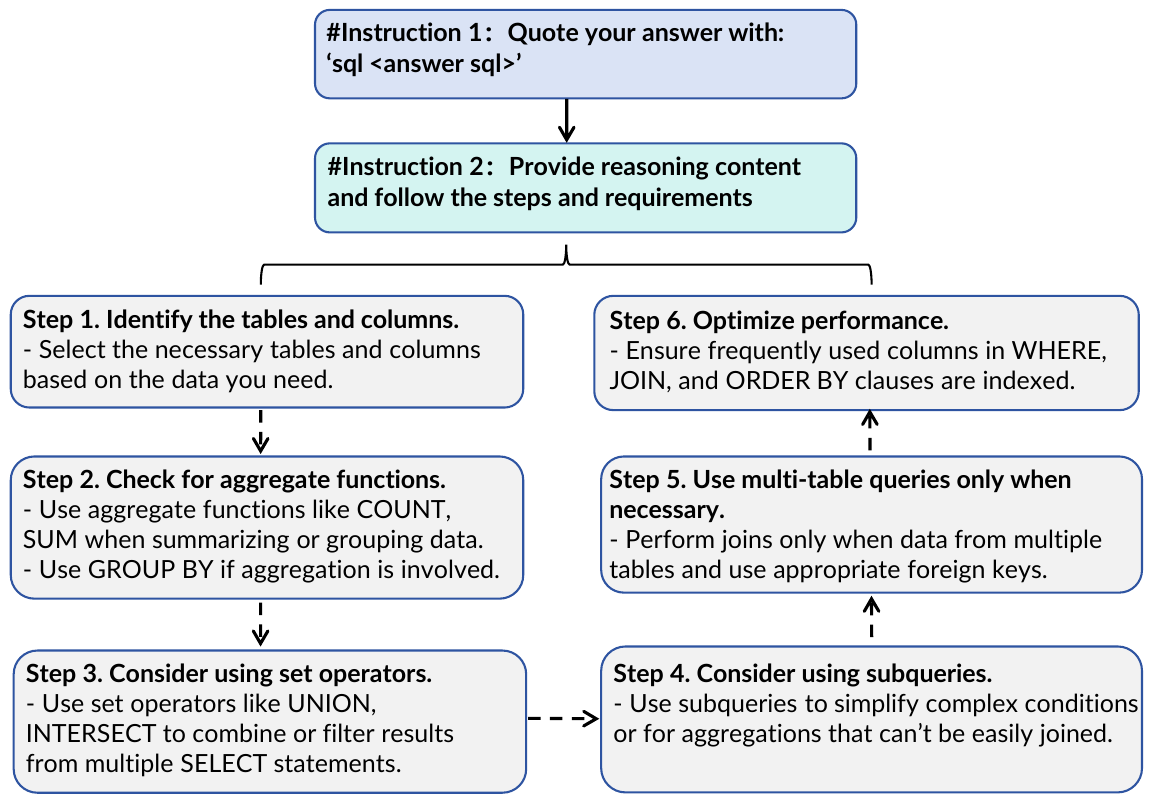}
  \caption{Reasoning instruction for the SQL generation.}
  \Description{}
  \label{fig:instru_sql_generation}
\end{figure}

To enhance the LLM's structured SQL generation capability, we incorporate the progressive reasoning process into the self‑enhanced fine‑tuning framework (Section~\ref{subsec:self_enhanced_ft}). In terms of data preparation, we collect reasoning-SQL pairs and apply filtering based on by SQL correctness to curate high-quality traces for fine-tuning. To align reasoning and generation, we provide role-specific prompt annotations (Figure~\ref{fig:instru_sql_generation}) and few-shot examples. The LLM outputs both the SQL statement and an accompanying stepwise rationale (as illustrated in Figure~\ref{fig:case_sql_gen}). This improves execution accuracy and interpretability, providing a transparent path from question to query.

\begin{figure}[t]
  \begin{tikzpicture}
  \node (table) [inner sep=0pt] {
  \begin{adjustbox}{width=0.9\linewidth}
    \begin{tabular}{l} 
     \\ [-2.0ex]
     \rowcolor{gray!10}\makecell[l]{
     \textbf{Input:} \\
     \emph{Schemas}: \textcolor{purple}{CREATE TABLE} ``singer" (``Singer\_ID" int, \\
     ``Name" text, ``Country" text, \textcolor{purple}{PRIMARY KEY} (``Singer\_ID");) \dots\\
     \emph{Question}: \textit{How many singers are from each country?} \\
     \\[-1.0ex]
     \textbf{LLM Output:} \\
    ```sql\\
    \textcolor{purple}{SELECT} singer.country , COUNT ( * ) \textcolor{purple}{FROM} singer \\
    \textcolor{purple}{GROUP BY} singer.country\\
    '''\\
    \textit{Here's the step-by-step reasoning:}\\
    \textit{1. Identify the necessary tables and columns: }\\
    \textit{We need to find the number of singers from each country. }\\
    \textit{The relevant table is `singer', and we need the `country' column.}\\
    \textit{2. Check for aggregate functions: }\\
    \textit{We need to count the number of singers from each country,}\\
    \textit{so we use the `COUNT(*)' aggregate function.}\\
    \textit{3. Consider using GROUP BY: }\\
    \textit{Since we want to group the results by country,}\\
    \textit{we use the `GROUP BY' clause.}\\
    \textit{\dots} \\ 
     }
  \end{tabular}
  \end{adjustbox}
  };
  \end{tikzpicture}
     \caption{Demonstration of the SQL generation.}
   \label{fig:case_sql_gen}
   \Description{}
\end{figure}

\subsection{Error-Aware Revision}
\label{subsec:error_aware_revision}

Despite the strong reasoning capabilities of LLMs, their generated SQL statements often contain structural or semantic flaws that hinder execution correctness. Empirical analysis reveals four dominant failure modes: \textit{(i)} invalid table-column relationships, \textit{(ii)} syntax violations such as misquoted identifiers, \textit{(iii)} misuse of unsupported operators or functions, and \textit{(iv)} field ambiguity due to alias conflicts or missing table qualifiers. These errors not only prevent successful execution but also obscure the model's reasoning trajectory.

To address this challenge, we propose an error-aware revision that transforms runtime errors into weak supervision signals for iterative SQL revision. Specifically, when a generated SQL query fails to execute, we extract the returned error message and integrate it into the CoT prompting framework, which is specifically tailored for debugging. This prompt explicitly instructs the model to \textit{(i)} analyze the error feedback, \textit{(ii)} identify the likely fault location, and \textit{(iii)} propose a minimal, logically consistent fix.

\begin{figure}[t]
  \centering
\includegraphics[width=0.9\linewidth]{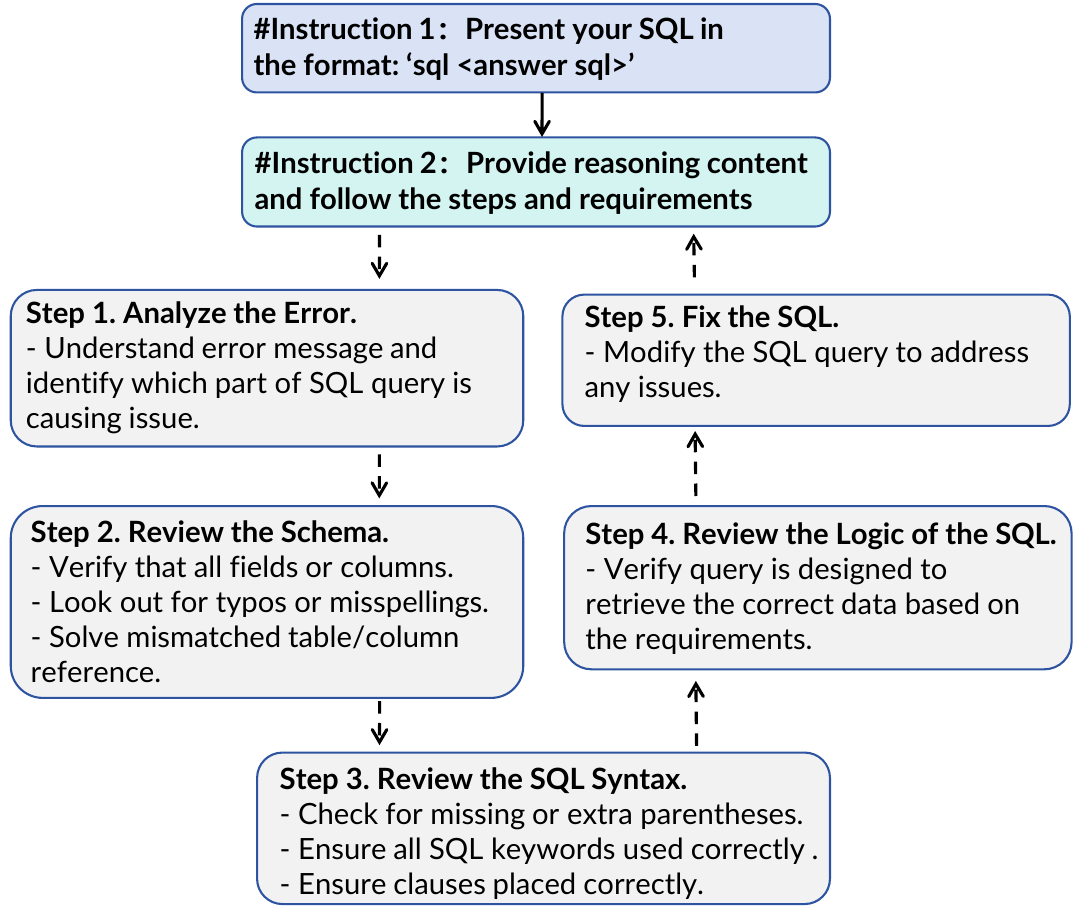}
  \caption{Reasoning instruction for the error-aware revision.}
  \Description{}
  \label{fig:instru_sql_correct}
\end{figure}

This mechanism enables \textbf{self-reflective reasoning}, where the model learns to revise both syntax and semantics based on runtime feedback. The correction process is structured in multiple stages, beginning with error diagnosis (e.g., resolving naming mismatches or unsupported syntax) and followed by logical reassessment of the modified query. 
Figure~\ref{fig:instru_sql_correct} illustrates our debugging instruction template, designed to ensure that each step is traceable and grounded in the original error context. A concrete example is shown in Figure~\ref{fig:case_sql_correct}, where an initial query fails due to a casing mismatch in column names (i.e., \texttt{pet\_type} vs. \texttt{PetType}). The model, guided by our CoT prompt, correctly localizes the issue and generates a revised query that is both syntactically valid and semantically faithful to the original intent.


\begin{figure}[t]
  \begin{tikzpicture}
  \node (table) [inner sep=0pt] {
  \begin{adjustbox}{width=0.9\linewidth}
    \begin{tabular}{l} 
     \\ [-2.0ex]
     \rowcolor{gray!10}\makecell[l]{
     \textbf{Input:} \\
     \emph{Schemas}: \textcolor{purple}{CREATE TABLE} Pets (PetID INTEGER, \\
     PetType VARCHAR(20), pet\_age INTEGER, weight REAL, \\
    \textcolor{purple}{PRIMARY KEY}  (PetID);) \dots\\ 
    \emph{Question}: \textit{What is the type and weight of the youngest pet?} \\
    \emph{Wrong sql}: \textcolor{purple}{SELECT} pets.pet\_type, pets.weight \textcolor{purple}{FROM} pets \\
    \textcolor{purple}{ORDER BY} pets.pet\_age ASC LIMIT 1\\
    \emph{Error information}: \textit{Error executing SQL} ``\textcolor{purple}{SELECT} pets.pet\_type, \\
    pets.weight \textcolor{purple}{FROM} pets \textcolor{purple}{ORDER BY} pets.pet\_age ASC LIMIT 1": \\
    ``\textit{no such column: pets.pet\_type}" \\
    \\ [-1.0ex]
     \textbf{LLM Output:} \\
    ```sql\\
    \textcolor{purple}{SELECT} pets.PetType, pets.weight \textcolor{purple}{FROM} pets \\
    \textcolor{purple}{ORDER BY} pets.pet\_age ASC LIMIT 1; \\
    '''\\
    \textit{\#\#\# Reasoning:} \\
    \textit{1.**Analyze the Error**:}  \\
    \textit{The error message indicates that there is no}  \\ 
    \textit{such column as `pet\_type' in the `pets' table.}  \\
    \textit{However, the correct column name is `PetType' (with a capital `T').} \\
    \textit{\dots} \\
     }
  \end{tabular}
  \end{adjustbox}
  };
  \end{tikzpicture}
       \caption{Demonstration of the error-aware revision.}
   \label{fig:case_sql_correct}
   \Description{}
\end{figure}

This revision module not only enhances robustness at inference time but also produces corrected samples that can be recycled into the training pipeline, enabling the model to internalize patterns of failure and repair. Over time, this improves generalization in complex, error-prone SQL generation scenarios.

\section{Experiments}

We conduct a comprehensive evaluation of the proposed \sysname framework for text-to-SQL to address the following questions:

\textbf{RQ1: Overall Performance.}  How does \sysname compare against state-of-the-art baselines? 

\textbf{RQ2: Ablation Study.} What is the individual and collective contribution of each \sysname module?

\textbf{RQ3: Parameter Sensitivity.} How do hyperparameters affect the \sysname's effectiveness?

\textbf{RQ4: Computation Efficiency.} What are the token efficiency and inference latency to produce SQL queries?

\textbf{RQ5: Human Evaluation.} How do human evaluators assess the quality of the generated SQL queries?

\subsection{Experimental Setup}
\subsubsection{Datasets}
We evaluate \sysname on two standard text-to-SQL benchmarks: Spider \cite{yu-etal-2018-spider} and Bird \cite{NEURIPS2023_83fc8fab}. 
\begin{itemize}
    \item \textbf{Spider}: This large-scale cross-domain benchmark contains 10,181 questions (including 1,034 validation samples) paired with 5,693 unique complex SQL queries across 200 databases. Each database features multiple tables spanning 138 diverse domains. Spider is particularly designed to assess a model's schema \textit{generalization capabilities} in generating syntactically and semantically correct SQL queries.
    \item \textbf{Bird}: Comprising 12,751 question-SQL pairs from 95 real-world databases, this benchmark covers 37+ specialized domains including healthcare, education, and blockchain. Bird introduces additional challenges by requiring systems to handle substantial database contents during SQL parsing, better reflecting \textit{practical deployment scenarios}.
\end{itemize}

\begin{table*}[t]
  \centering
  \caption{Comparison of \sysname and baselines on the Spider and Bird datasets by Execution Accuracy (EX) and Valid Efficiency Score (VES). ``Direct'' denotes the zero-shot prompt baseline. Best and second-best results are \textbf{bolded} and \underline{underlined}, respectively. 
  The GPT-4 results on Spider and Bird are taken from \cite{DBLP:journals/pvldb/GaoWLSQDZ24} and \cite{wang-etal-2025-mac}, respectively. As full-parameter training exceeds our hardware limits, MAC-SQL results are adopted from \cite{wang-etal-2025-mac}. All others are from our implementations.
  \label{tab:main_results}}

  \scalebox{0.88}{
  \begin{tabular}{l l l c c c c}
    \toprule
    \textbf{Category} & \textbf{Method} & \textbf{Model} & \textbf{EX (Spider)} & \textbf{VES (Spider)} & \textbf{EX (Bird)} & \textbf{VES (Bird)} \\
    \midrule
    \multirow{5}{*}{\textit{Direct}} 
    & & GPT-4 & 72.30 & - & 46.35 & 49.77 \\
    & & Qwen2.5-7B & 75.30 & 71.67 & 45.44 & 48.37 \\
    & & Llama-3.1-8B & 56.00 & 53.43 & 41.20 & 46.58 \\
    & & DeepSeek-R1-Qwen-7B ~\cite{guo2025deepseek} & 39.10 & 12.18 & 6.17 & 7.05\\
    & & DeepSeek-R1-Llama-8B ~\cite{guo2025deepseek} & 41.50 & 18.93 & 15.12 & 15.98 \\
    \midrule
    \multirow{9}{*}{\textit{Shallow prompt-based methods}}
    & \multirow{2}{*}{PURPLE}
    & Qwen2.5-7B &\textbf{81.90} & \textbf{80.97} & 36.31 & 42.33\\
    & &Llama-3.1-8B &70.60& 73.03 & 28.49& - \\
    \cmidrule(lr){2-7}
    & \multirow{2}{*}{MetaSQL}
    &Qwen2.5-7B & 68.60 & 64.98& -& -\\
    & &Llama-3.1-8B &65.40& 62.18& -& - \\
    \cmidrule(lr){2-7}
    & \multirow{2}{*}{DAIL-SQL}
    & Qwen2.5-7B & 72.40 & 67.71 & 39.90 & 41.59 \\
    & &Llama-3.1-8B & 74.80 & 35.05 & 41.00 & 43.82 \\
    \cmidrule(lr){2-7}
    & \multirow{2}{*}{DAC}
    & Qwen2.5-7B & 76.80 & 72.95 & 46.54 & 48.25 \\
    & &Llama-3.1-8B & 75.70 & 71.93 & 45.57 & 48.32 \\
    \midrule
    \multirow{3}{*}{\textit{CoT-based methods}}
    & \multirow{2}{*}{DEA-SQL}
    & Qwen2.5-7B & 52.10 & 49.52 & 13.17 & 14.93 \\
    & &Llama-3.1-8B & 19.00 & 16.76 & 8.41 & 8.35 \\ 
    
    \cmidrule(lr){2-7}
    & \multirow{2}{*}{DIN-SQL}
    & Qwen2.5-7B & 41.00 & 40.01 & 39.96 & 41.41 \\
    & & Llama-3.1-8B & 55.30 & 55.48 & 38.72 & 40.65 \\
    \midrule
    
    \multirow{3}{*}{\textit{Fine-tuned-based methods}} 
    &MAC-SQL & SQL-Llama (7B)~\cite{wang-etal-2025-mac} & 76.25 & - & 43.94 & \underline{57.36} \\
    \cmidrule(lr){2-7}
    & \multirow{2}{*}{DTS-SQL} & Qwen2.5-7B ~\cite{yang2024qwen2} & 73.50 & 68.96 & 42.18 & 44.43 \\
    & & Llama-3.1-8B ~\cite{dubey2024llama} & 71.00 & 66.93 & 41.92 & 43.59 \\
    \midrule
    & \multirow{2}{*}{\textbf{\texttt{CoTE-SQL (ours)}}}
     & Qwen2.5-7B & 77.80 & 74.68 & \textbf{53.39} & \textbf{59.02} \\
     & & Llama-3.1-8B & \underline{79.60} & \underline{77.19} & \underline{49.54} & 55.25 \\
    \bottomrule
  \end{tabular}
  }
\end{table*}

\subsubsection{Evaluation Metrics}

We use two evaluation metrics: Execution Accuracy (EX) and Valid Efficiency Score (VES), following the official evaluation scripts of Spider and Bird with minor adaptations.

\begin{itemize}
    \item \textbf{Execution Accuracy (EX) ~\cite{yu-etal-2018-spider}}: Measures the proportion of cases where the execution result of the predicted SQL query matches that of the ground truth. This metric reflects the correctness of model outputs in terms of execution semantics.

    \item \textbf{Valid Efficiency Score (VES) ~\cite{NEURIPS2023_83fc8fab}}: Assesses both the correctness and efficiency of valid SQL queries generated by the model. A predicted SQL is considered \textit{valid} if its execution result aligns with that of the ground-truth query. The efficiency aspect captures factors such as execution time, throughput, memory consumption, or a composite of these metrics.
\end{itemize}

\subsubsection{Baselines}

We compare our approach with several state-of-the-art text-to-SQL baselines, which fall into two main categories: (i) fine-tuning-based methods~\cite{pourreza-rafiei-2024-dts, wang-etal-2025-mac}, and (ii) prompt-based methods~\cite{NEURIPS2023_72223cc6, DBLP:journals/pvldb/GaoWLSQDZ24, wang2024dac,xie-etal-2024-decomposition,fan2024metasql, ren2024purple}.
For \textit{fine-tuning-based methods}, we either reproduce the results using their official training procedures or cite the results reported in the original papers when reproduction is infeasible. For instance, MAC-SQL~\cite{wang-etal-2025-mac} requires full-parameter fine-tuning, which could not be executed due to out-of-memory issues on our hardware. In such cases, we directly report the performance from the original work. All other methods in this category were trained and evaluated by us.
For \textit{prompt-based methods}, we report the zero-shot performance of both open-source and closed-source LLMs, as well as results from adapting open-source models to existing text-to-SQL frameworks. GPT-4 results are taken from prior work~\cite{wang-etal-2025-mac}, while all other results are produced through our experiments. The baselines evaluated are as follows:

\begin{itemize}
  \item \textbf{DTS-SQL}~\cite{pourreza-rafiei-2024-dts}: A two-stage fine-tuning method that decomposes the task to improve the LLMs' performance.
  \item \textbf{MAC-SQL}~\cite{wang-etal-2025-mac}: A full-parameter fine-tuned agent framework for complex, multi-step reasoning in text-to-SQL tasks.
  \item \textbf{DIN-SQL}~\cite{NEURIPS2023_72223cc6}: A decomposition-based method that divides complex tasks into sub-problems using contextual prompts.
  \item \textbf{MetaSQL}~\cite{fan2024metasql}: Converts queries into metadata for guiding SQL generation through constrained prompting. Evaluated only on the Spider dataset due to incompatibility with Bird.
  \item \textbf{PURPLE}~\cite{ren2024purple}: Uses pre-trained models to retrieve few-shots with key logical structures for effective SQL generation.
  \item \textbf{DAIL-SQL}~\cite{DBLP:journals/pvldb/GaoWLSQDZ24}: Assesses various demonstration selection strategies to enhance few-shot text-to-SQL performance.
  \item \textbf{DAC}~\cite{wang2024dac}: Introduces decomposed automatic correction mechanisms to refine SQL decoding.
  \item \textbf{DEA-SQL}~\cite{xie-etal-2024-decomposition}: Employs query decomposition workflows to improve LLMs' attention and reasoning on complex queries.
\end{itemize}


\subsubsection{Implementation Details of \sysname}
We employ the Llama-3.1-8B model, aligned via instruction tuning~\cite{dubey2024llama}, as the base model. Both the \textit{Schema Selection} and \textit{SQL Generation} modules follow a 3-shot inference strategy. Demonstration examples are retrieved using the BM25 algorithm~\cite{INR-019}, which selects the most relevant samples from a candidate pool for each user query. For the Spider dataset, we utilize the entity-linking annotations provided by~\cite{liu-etal-2021-awakening} as the retrieval corpus. As the Bird dataset lacks such annotations, the \textit{Schema Selection} module is trained exclusively on the Spider dataset.


For fine-tuning, we apply Low-Rank Adaptation (LoRA)~\cite{DBLP:conf/iclr/HuSWALWWC22} to the base model using 3,000 training examples randomly sampled from the Spider and Bird training sets (1,500 each). Training is conducted with the Transformers library (v4.45.2) on a single NVIDIA A100-SXM4-80GB GPU. The LoRA configuration uses a rank of 16 and an alpha value of 16. We train with bfloat16 precision, a learning rate of $5 \times 10^{-5}$, for 5 epochs, using a batch size of 1 and gradient accumulation over 4 steps. Total training time is approximately 3 GPU hours.
Inference is performed using consistent generation settings across all modules: a temperature of 0.2, a maximum sequence length of 2048 tokens, top-p set to 1.0, and a single output per query. All experiments use a fixed random seed of 42 for reproducibility.


\subsection{Overall Performance (RQ1)}

We conduct a comprehensive comparison with existing baselines, and the results are reported in Table~\ref{tab:main_results}. On the \textit{Bird} dataset, \sysname achieves state-of-the-art performance, surpassing all baselines by 6.85\%-47.22\% in EX and by 1.66–51.97 in VES. On the \textit{Spider} dataset, \sysname outperforms all baselines except PURPLE with Qwen2.5-7B, achieving improvements of 2.8\%–60.6\% in EX and 4.24–65.01 in VES over the rest. The PURPLE method yields the best results with Qwen2.5-7B on the Spider dataset, as it has been carefully fine-tuned on a skeleton prediction model for this dataset.
Results show that \sysname demonstrates strong generalization across datasets with varying schema complexity and domain coverage.
We further analyze the experimental results and summarize several key observations below.


\begin{figure}[t]
    \centering
    \begin{subfigure}{0.9\linewidth}
        \captionsetup{skip=0pt}
        \includegraphics[width=\linewidth]{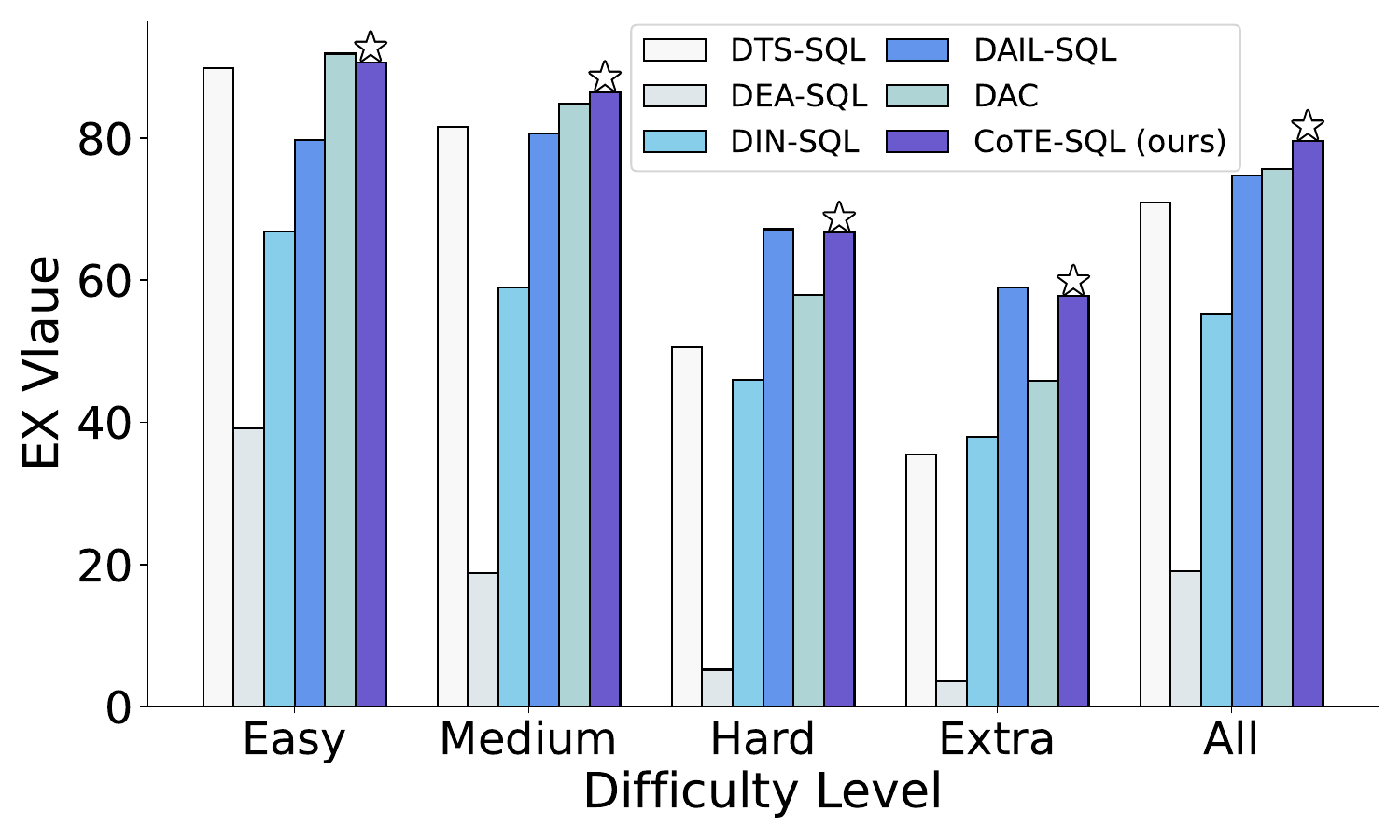}
        \caption{Spider dataset.}
        \label{fig:diff_ex_spider}
    \end{subfigure}
    \begin{subfigure}{0.85\linewidth}
        \captionsetup{skip=0pt}
        \includegraphics[width=\linewidth]{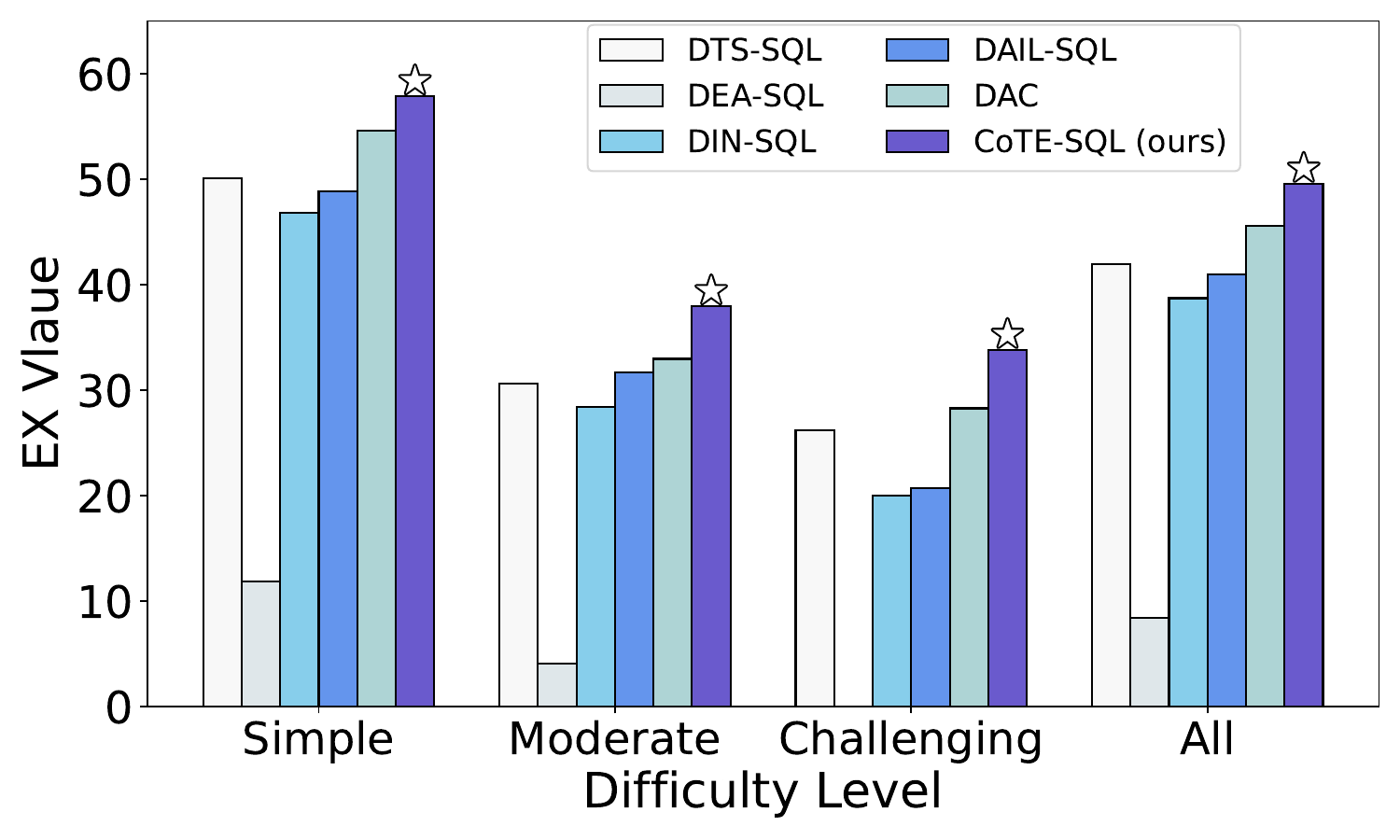}
        \caption{Bird dataset.}
        \label{fig:diff_ex_bird}
    \end{subfigure}
    \caption{Comparison of EX performance across different difficulty levels for two datasets.
    }
    \label{fig:diff_2_images}
    \Description{}
\end{figure}

Compared with direct inference using LLMs, \sysname decomposes the SQL generation process into fine-grained sub-tasks, allowing the model to focus on specific aspects of the task, leading to more accurate and efficient SQL generation. This structured generation pipeline significantly improves overall performance.
While methods like DIN-SQL and DEA-SQL also adopt task decomposition strategies, their designs are often overly fragmented and heavily rely on template-based CoT prompts that supply predefined, concrete reasoning steps in the few-shot examples. These prompts are challenging for LLMs, especially smaller ones, to interpret and generalize effectively, which results in degraded performance. In fact, under small-scale LLM settings, these methods can underperform even simple direct inference with large-scale LLMs, e.g., GPT-4. In contrast, \sysname’s flexible CoT prompting strategy enables adaptive reasoning paths, resulting in strong generalization across queries with varying structural and reasoning complexity.


Notably, although DeepSeek-R1-Qwen-7B and DeepSeek-R1-Llama-8B inherit some reasoning capabilities from the teacher model DeepSeek-R1, they still perform poorly on text-to-SQL tasks. We find that their generated CoT reasoning is often disorganized and frequently fails to produce complete SQL statements. This can be attributed to the lack of high-quality, domain-specific CoT data during the distillation process, particularly in the context of the highly specialized text-to-SQL task. These results further validate the effectiveness of \sysname in enhancing the reasoning capability and generation quality of small-scale LLMs.


In contrast to end-to-end fine-tuning methods like DTS-SQL and MAC-SQL, \sysname adopts a self-enhanced reasoning approach, where the LLM is lightly fine-tuned using fundamental principle data generated by itself. This improves the model's understanding of the text-to-SQL task and boosts its reasoning ability. It is worth emphasizing that, while MAC-SQL relies on full-parameter fine-tuning, \sysname achieves superior performance with only partial parameter updates, demonstrating both effectiveness and parameter efficiency.


Furthermore, we profile performance across different difficulty levels, as shown in Figure~\ref{fig:diff_2_images}. On the Spider dataset, \sysname achieves consistent improvements over the strongest baseline across medium, hard, extra, and all difficulty levels, with gains of 1.7\%, 8.7\%, 12\%, and 3.9\%, respectively. On the Bird dataset, the improvements are 3.25\%, 4.96\%, 5.51\%, and 3.97\% on simple, moderate, challenging, and all levels, respectively.These consistent improvements across difficulty levels demonstrate that \sysname generalizes well to queries of varying complexity across diverse domains.
Notably, the most significant improvements are observed on the most challenging levels. This is primarily due to the structured reasoning guided by CoT prompts, which help the model construct more transparent logical chains, particularly beneficial in complex scenarios, thus significantly improving the quality and accuracy of generated SQL statements.

\begin{table*}[t]
  \centering
  \caption{Migration results of DAC and PURPLE to newer and stronger models. \sysname's results are included for comparison. Best and second-best results are \textbf{bolded} and \underline{underlined}.}
  \label{tab:migration_results}
  \scalebox{0.85}{
  \begin{tabular}{l l c c c c}
    \toprule
    \textbf{Method} & \textbf{Model} & \textbf{EX (Spider)} & \textbf{VES (Spider)} & \textbf{EX (Bird)} & \textbf{VES (Bird)} \\
    \midrule
    \multirow{2}{*}{DAC} 
    & Qwen3.5-9B        & 75.70 & 73.48 & \underline{51.31} & \textbf{59.08} \\
    & IQuest-Coder-V1-7B & 71.60 & 69.72 & 36.77 & 43.33 \\
    \cmidrule(lr){1-6}
    \multirow{2}{*}{PURPLE}
    & Qwen3.5-9B        & 77.40 & \underline{76.76} & 34.49 & 39.19 \\
    & IQuest-Coder-V1-7B & 50.00 & 50.68 & 14.54 & 18.01 \\
    \midrule
    \multirow{2}{*}{\textbf{\sysname (ours)}}
    & Qwen2.5-7B        & \underline{77.80} & 74.68 & \textbf{53.39} & \underline{59.02} \\
    & Llama-3.1-8B      & \textbf{79.60} & \textbf{77.19} & 49.54 & 55.25 \\
    \bottomrule
  \end{tabular}
  }
\end{table*}

The baseline methods in Table~\ref{tab:main_results} are built on strong open-source models at the time. A natural question is whether \sysname would still remain effective if these methods are migrated to stronger models with improved logical and coding abilities. To investigate this, we re-evaluate the two best baselines, DAC and PURPLE, by migrating them to Qwen3.5-9B and IQuest-Coder-V1-7B~\cite{yang2026iquest}. As shown in Table~\ref{tab:migration_results}, PURPLE exhibits a performance decline after migration. For the Bird dataset, DAC improves after migrating to Qwen3.5-9B, with VES slightly surpassing \sysname (by 0.06) due to the model’s preference for more execution-efficient SQL. However, \sysname still outperforms on the core EX metric. This indicates that \sysname through self-enhanced fine-tuning and structured reasoning, possesses a fundamental advantage that surpasses model iterations.

\subsection{Ablation Study (RQ2)}
We conduct an ablation study by individually removing the \textit{Schema Selection}, \textit{SQL Correction}, and \textit{Self-Enhanced Reasoning} modules from \sysname. Table~\ref{tab:Ablation_Study_Spider_All} reports the results of these variants on both the Spider and Bird datasets. \sysname consistently outperforms all its ablated variants, demonstrating the effectiveness of each module. Specifically, the \textit{Self-Enhanced Reasoning} module enhances the reasoning ability by fine-tuning on self-generated CoT data, leading to improvements of 4.30\% on Spider and 1.30\% on Bird. The \textit{Schema Selection} module filters out irrelevant schema elements, allowing the model to focus on key components and CoT instructions, resulting in gains of 2\% on Spider and 1.63\% on Bird. Lastly, the \textit{SQL Correction} module reduces syntax and logical errors in generated SQL, improving execution accuracy by 1.60\% on Spider and 5.67\% on Bird.


Due to the differing characteristics of the datasets, the impact of each module varies. The Spider dataset emphasizes generalization across diverse schemas, where improvements in reasoning capabilities yield larger gains. Conversely, the Bird dataset features substantially larger databases, where syntax errors and incorrect table-field associations are more frequent, thus highlighting the importance of SQL correction. Overall, the ablation study confirms that all modules contribute significantly to the strong performance of \sysname.


\begin{table}[t]
  \centering
  \caption{Ablation results. Numbers in parentheses denote the decrease in EX performance when individual components are ablated from \sysname.\label{tab:Ablation_Study_Spider_All}}

  
  \begin{tabular}{l c c}
    \toprule
    \textbf{Method} & \textbf{EX (Spider)} & \textbf{EX (Bird)} \\
    \midrule
    w/o Self-Enhanced Fine-Tuning & 75.30 \textcolor{gray}{(-4.30)} & 48.24 \textcolor{gray}{(-1.30)} \\
    w/o Schema Selection & 77.60 \textcolor{gray}{(-2.00)} & 47.91 \textcolor{gray}{(-1.63)} \\
    w/o SQL Correction  & 78.00 \textcolor{gray}{(-1.60)} & 43.87 \textcolor{gray}{(-5.67)} \\
    \textbf{\sysname} & \textbf{79.60} & \textbf{49.54} \\
    \bottomrule
  \end{tabular}

\end{table}

We further profile the ablation results across different difficulty levels. Table~\ref{tab:Ablation_Study_diff_Spider_DIFF} shows that removing the \textit{Schema Selection} module causes an EX drop of up to 7.20\% on the Extra difficulty level of the Spider dataset. This is attributable to the increased complexity and abundance of irrelevant schema elements in harder SQL tasks, which complicates reasoning and disperses model attention when schema filtering is absent. Meanwhile, removing the \textit{SQL Correction} module leads to significant performance declines (5.39\%-8.27\%) at all difficulty levels on the Bird dataset, as shown in Table~\ref{tab:Ablation_Study_diff_Bird}, underscoring its critical role in maintaining system effectiveness.


\begin{table*}[t]
  \centering
  \caption{EX performance across questions with different difficulty levels in the ablation study on the Bird dataset.\label{tab:Ablation_Study_diff_Bird}}

  \begin{tabular}{l c c c c}
    \toprule
    \textbf{Method} & \textbf{Simple} & \textbf{Moderate} & \textbf{Challenging} & \textbf{All} \\
    \midrule
    w/o Self-Enhanced Fine-Tuning & 57.30 \textcolor{gray}{(-0.54)} & 36.42 \textcolor{gray}{(-1.51)} & 28.28 \textcolor{gray}{(-5.51)} & 48.24 \textcolor{gray}{(-1.30)} \\
    
    w/o Schema Selection  & 56.54 \textcolor{gray}{(-1.30)} & 36.85 \textcolor{gray}{(-1.08)} & 28.28 \textcolor{gray}{(-5.51)} & 47.91 \textcolor{gray}{(-1.63)} \\
    
    w/o Correction  & 52.43 \textcolor{gray}{(-5.41)} & 32.54 \textcolor{gray}{(-5.39)} & 25.52 \textcolor{gray}{(-8.27)} & 43.87 \textcolor{gray}{(-5.67)} \\
     
    \textbf{\sysname}   & \textbf{57.84} & \textbf{37.93} & \textbf{33.79} & \textbf{49.54} \\
    \bottomrule
  \end{tabular}
\end{table*}

\begin{table*}[t]
  \centering
  \caption{EX performance across questions with different difficulty levels in the ablation study on the Spider dataset.\label{tab:Ablation_Study_diff_Spider_DIFF}}
  
  \begin{tabular}{l c c c c c}
    \toprule
    \textbf{Method} & \textbf{Easy} & \textbf{Medium} & \textbf{Hard} & \textbf{Extra} & \textbf{All} \\
    \midrule
    w/o Self-Enhanced Fine-Tuning & 87.10 \textcolor{gray}{(3.60)} & 80.90 \textcolor{gray}{(-5.60)} & 64.90 \textcolor{gray}{(-1.80)} & 53.60 \textcolor{gray}{(-4.20)} & 75.30 \textcolor{gray}{(-4.30)} \\
    w/o Schema Selection  & 89.50 \textcolor{gray}{(-1.20)} & 86.10 \textcolor{gray}{(-0.40)} & 64.40 \textcolor{gray}{(-2.30)} & 50.60 \textcolor{gray}{(-7.20)} & 77.60 \textcolor{gray}{(-2.00)}  \\
    w/o Correction  & 89.10 \textcolor{gray}{(-1.60)} & 84.50 \textcolor{gray}{(-2.00)} & 64.90 \textcolor{gray}{(-1.80)} & 57.80 \textcolor{gray}{(0.00)} & 78.00 \textcolor{gray}{(-1.60)} \\
    \textbf{\sysname}  & 90.70 & 86.50 & 66.70 & 57.80 & 79.60 \\
    \bottomrule
  \end{tabular}
  
\end{table*}

\begin{figure}[t]
    \centering
    \begin{subfigure}{0.38\linewidth}
        \captionsetup{skip=0pt}
        \includegraphics[width=\linewidth]{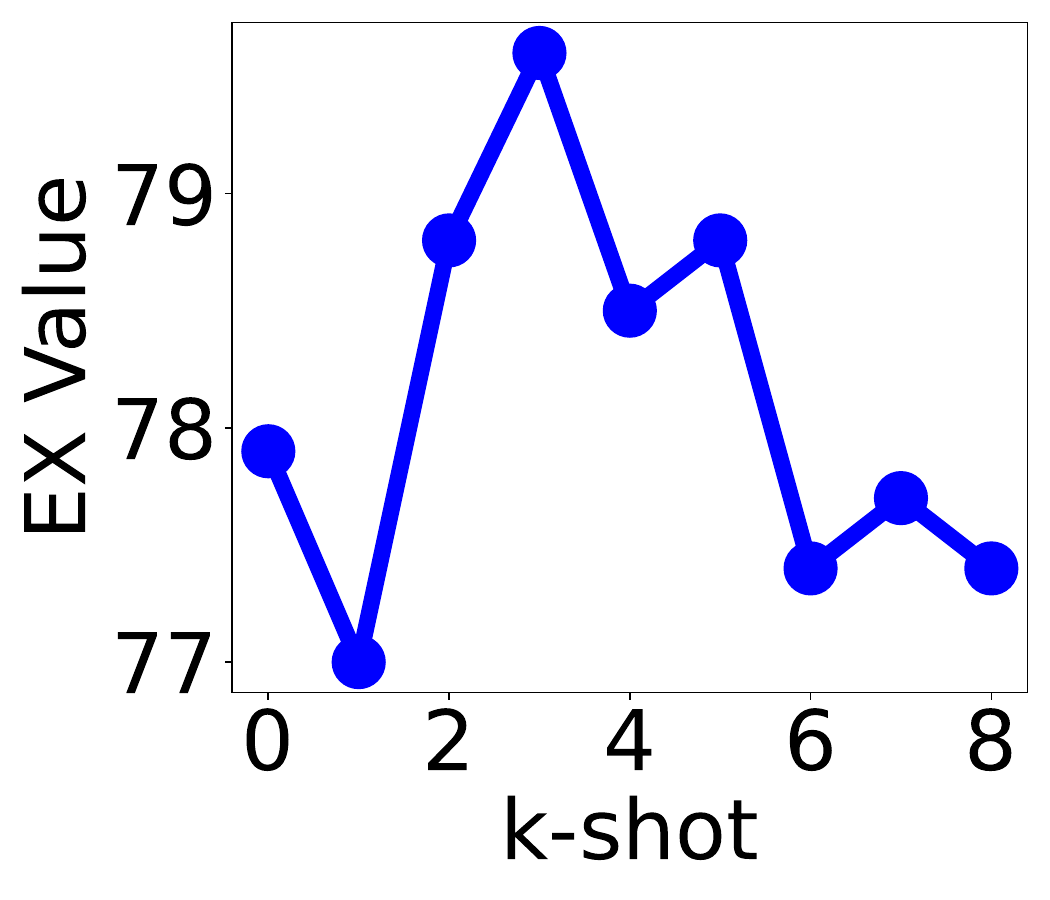}
        \caption{Spider dataset.}
        \label{fig:k-shot_spider}
    \end{subfigure}
    \begin{subfigure}{0.38\linewidth}
        \captionsetup{skip=0pt}
        \includegraphics[width=\linewidth]{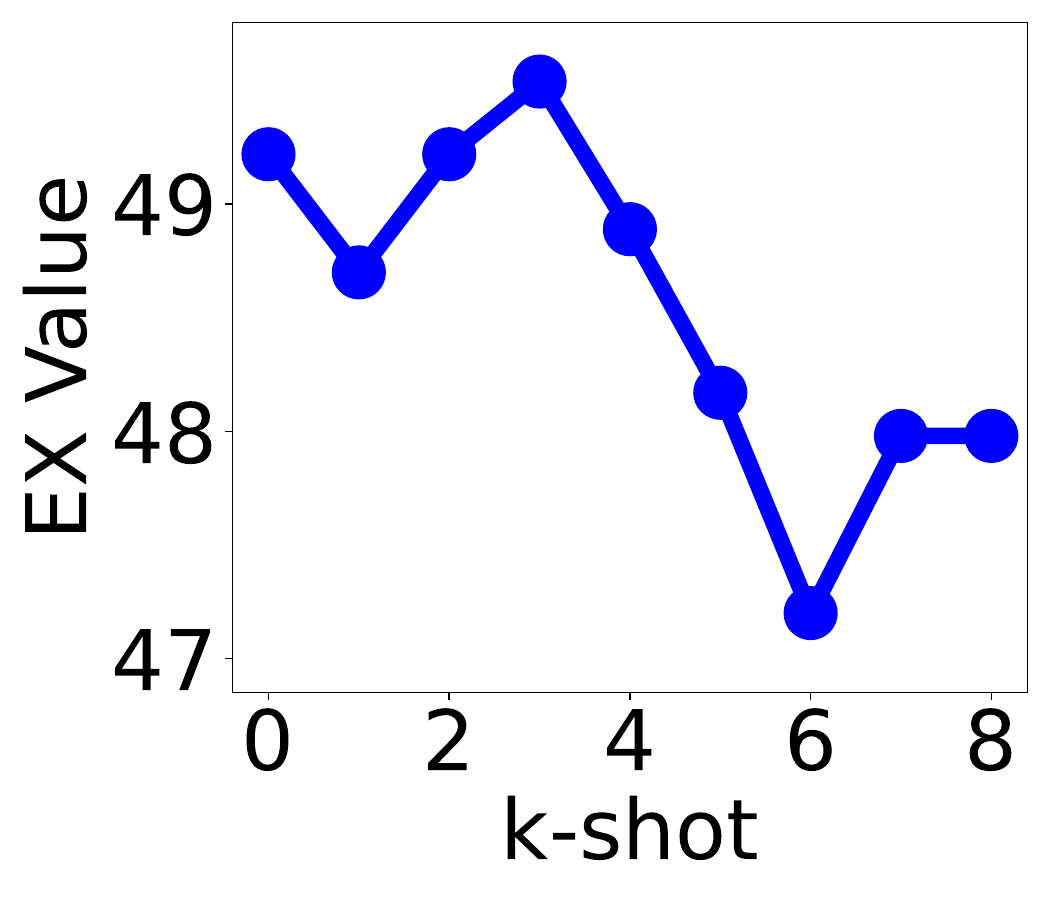}
        \caption{Bird dataset.}
        \label{fig:k-shot_bird}
    \end{subfigure}
    \caption{Impact of the number of few-shot samples. \label{fig:parameters-fewshot}}
    \Description{}
\end{figure}

\begin{figure}[t]
    \centering
    \begin{subfigure}{0.38\linewidth}
        \captionsetup{skip=0pt}
        \includegraphics[width=\linewidth]{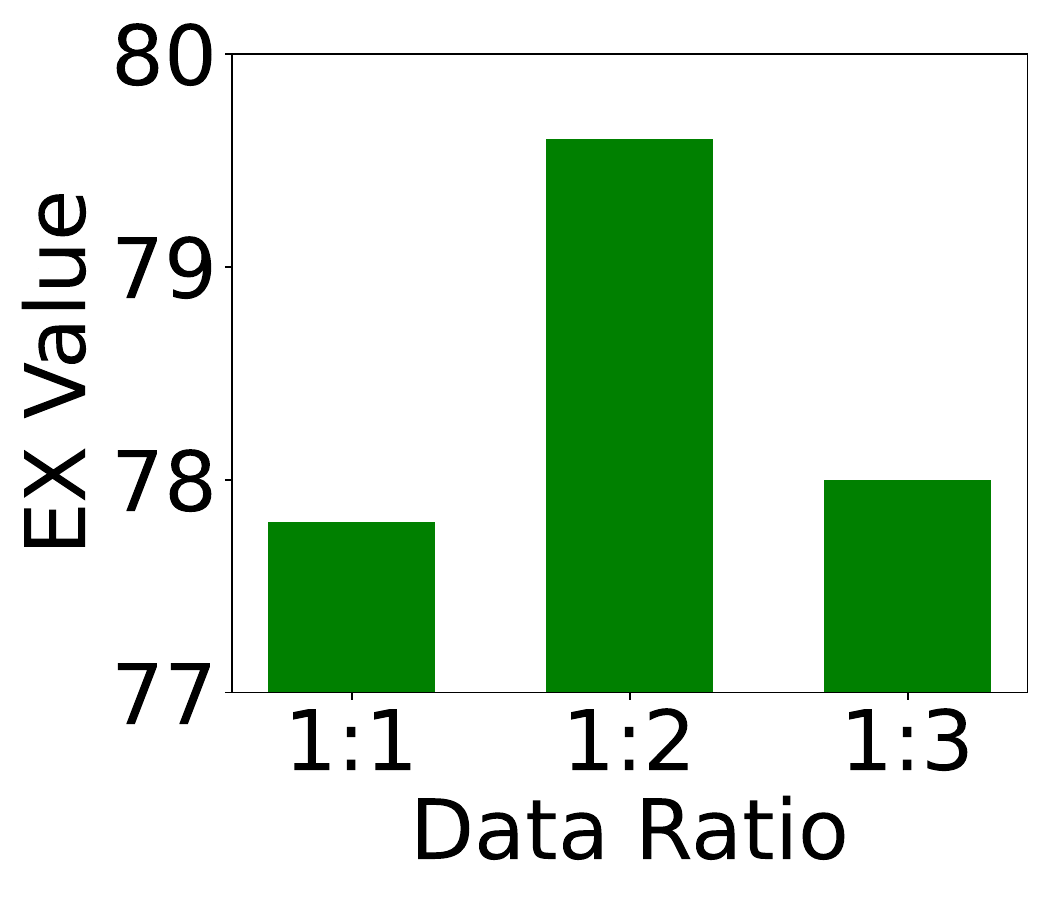}
        \caption{Spider dataset.}
        \label{fig:data-ratio_spider}
    \end{subfigure}
    \begin{subfigure}{0.38\linewidth}
        \captionsetup{skip=0pt}
        \includegraphics[width=\linewidth]{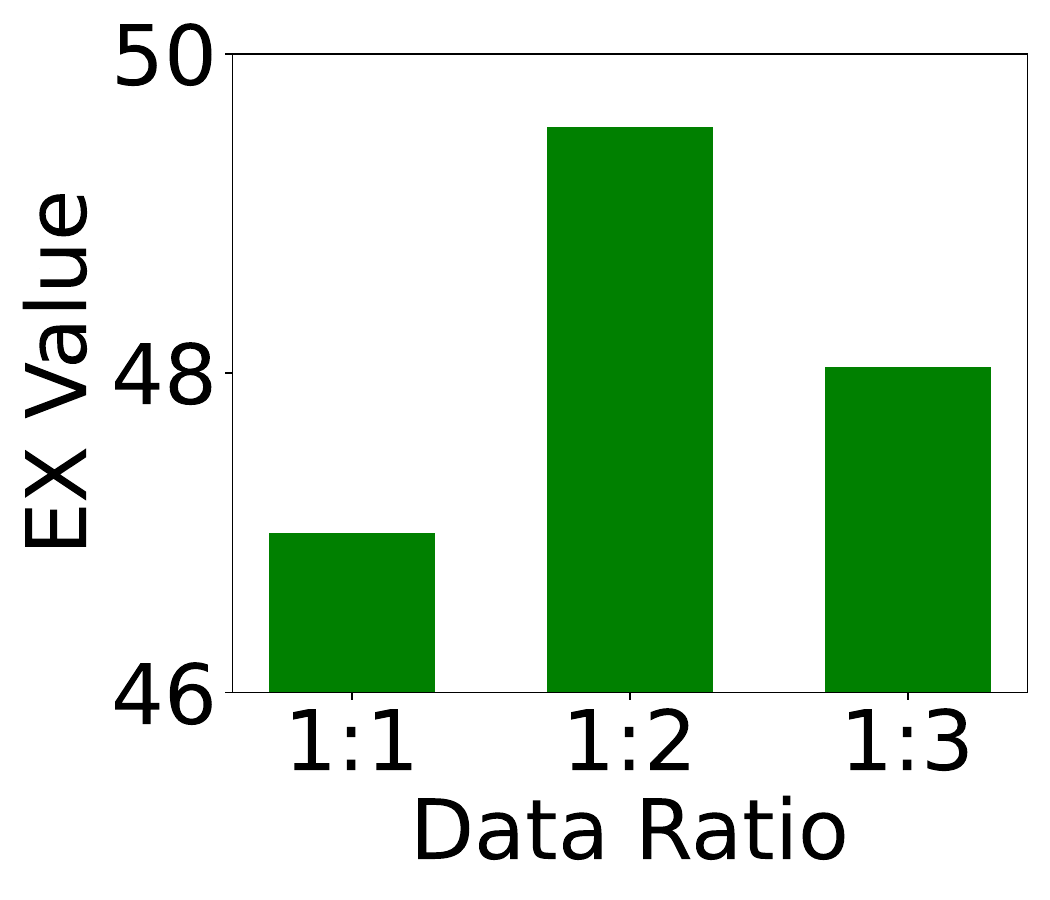}
        \caption{Bird dataset.}
        \label{fig:data-ratio_bird}
    \end{subfigure}
    \caption{
    Impact of training data ratio between schema selection and SQL generation during self-enhanced fine-tuning.\label{fig:parameters-dataratio}}
    \Description{}
\end{figure}

\subsection{Parameters Sensitivity (RQ3)}

We investigate the sensitivity of \sysname to two key parameters: the number of few-shot examples $k$ and the training data ratio between the schema selection and SQL generation tasks.
As shown in Figure~\ref{fig:parameters-fewshot}, the best performance is achieved when $k=3$, indicating that a moderate number of few-shot examples enhances open-source LLMs' understanding of the text-to-SQL task. When $k$ is too small, the selected examples may fail to capture the logical structure of the current SQL query, providing limited guidance. Conversely, when $k$ is too large, the excessive context introduces redundancy and distracts the model's attention, thereby degrading reasoning performance. 

Surprisingly, performance at $k=1$ is even lower than at $k=0$. A closer analysis of 100 samples under the $k=1$ setting reveals that in approximately 70\% of cases, the selected example, while lexically similar to the input question, follows a different SQL logic. This leads the model to overfit to an irrelevant pattern, thus impairing generalization.


Figure~\ref{fig:parameters-dataratio} presents results under varying training data ratios. A 1:2 ratio between schema selection and SQL generation yields the best overall performance. This reflects the relative difficulty of SQL generation, which requires producing syntactically correct and semantically meaningful queries that may involve joins, nested queries, and complex conditions, necessitating more extensive supervision compared to schema selection.


\begin{table}[t]
  \centering
  \caption{
  Token efficiency of different methods on Spider.\label{tab:Token Efficiency}}
\label{tab:token_efficiency}
  \begin{tabular}{c c c c }
    \toprule
    \textbf{Method} & \textbf{Avg. Token Num} & \textbf{Avg. Time} & \textbf{EX} \\
    \midrule
    DTS-SQL& \underline{1081} & 7.7s & 71.0\\ 
    DEA-SQL & 12,038 & 100.2s & 19.0\\
    DIN-SQL & 10,453 & 16.2s & 55.3\\
    DAIL-SQL &  \textbf{932} &  \textbf{2.4s} & 74.8\\
    DAC & 11,741 & \underline{7.4s} & \underline{75.7} \\
    \midrule
    \textbf{\sysname} & 2,242 & 10.9s & \textbf{79.6}\\
    \bottomrule
  \end{tabular}
\end{table}

\subsection{Computation Efficiency (RQ4)}



We evaluate average token count and inference time on 10 randomly sampled Spider queries. As shown in Table~\ref{tab:Token Efficiency}, \sysname uses significantly fewer tokens than multi-stage baselines (DEA-SQL, DIN-SQL, DAC), which incur redundant prompting and fragmented reasoning. For instance, DEA-SQL averages 12,038 tokens and over 100 seconds per query with 19.0\% EX, whereas \sysname requires only 2,242 tokens and 10.9 seconds, achieving 79.6\% EX.
This efficiency arises from our structured CoT prompt, which unifies reasoning and generation in a single-stage pipeline, avoiding redundant decomposition and repeated LLM calls.

When compared to DTS-SQL and DAIL-SQL, \sysname achieves superior accuracy (79.6\% vs. 74.8\% and 71.0\%, respectively), despite a moderate increase in token usage and inference time. Notably, DAIL-SQL achieves the lowest average token count (932) and inference time (2.4s), but this comes at the cost of reduced reasoning capacity, leading to suboptimal EX. Our method introduces minimal overhead by incorporating lightweight intermediate reasoning steps, which are crucial for handling complex and compositional SQL structures.


Moreover, while DAC achieves comparable execution accuracy (75.7\%) with relatively fast inference (7.4s), it requires over 11,000 tokens on average, suggesting that it trades off efficiency in token space for runtime gains via aggressive parallelization or optimization strategies. In contrast, \sysname offers a balanced trade-off between computational cost and predictive accuracy, making it more suitable for deployment in real-time systems with constrained budgets.



In summary, \sysname achieves the best accuracy-efficiency trade-off, outperforming multi-stage and end-to-end approaches, validating our compact, interpretable reasoning design that enhances both efficiency and semantic correctness.

\subsection{Human Evaluation (RQ5)} 



To assess generated SQL quality, we randomly sample 100 Spider questions and compared outputs from \sysname and the strongest baseline, DAC. We recruited three skilled volunteers with SQL knowledge, provided them with detailed evaluation criteria and background information. Evaluators saw each question, its gold SQL, and two anonymized outputs (order shuffled). Evaluation used three criteria: \textbf{Completeness (CP)}: Does the SQL include all necessary and relevant tables and fields required to answer the question? \textbf{Structural Soundness (SS)}: Does the SQL follow a clean and efficient structure with minimal redundancy? \textbf{Logical Consistency (LC)}: Which SQL better captures the question's logic, e.g., computation or condition logic? The preferred SQL received 1 point per criterion, aggregated over 100 samples and averaged among volunteers to obtain the final results.

As shown in Table~\ref{tab:human_eval}, \sysname outperforms DAC on all three criteria: 93 (CP), 83 (SS), and 86 (LC), exceeding DAC by 8, 3, and 7 points. EX and VES metrics are also reported, further validating our method.

Overall, these results suggest more reliable handling of diverse and complex SQL queries.


We present a case study in Figure~\ref{fig:case_human_eval} comparing SQL outputs from CoTE-SQL and DAC (both Llama-3.1-8B) for a specific question. CoTE-SQL correctly identifies target entities, applies joins, and produces a query equivalent to the ground truth. DAC omits the \textit{country} condition and fails to capture necessary relationships, reflecting limitations in its reasoning and schema understanding. This example highlights \sysname's advantage across all three metrics.

\begin{table}[t]
    \centering
    \caption{
    Comparison of human preference scores and quantitative metrics. The first three columns show human evaluations; the last two show corresponding quantitative metrics.\label{tab:human_eval}}
    \begin{tabular}{c | c c c | c c}
    \toprule
         \textbf{Method} & \textbf{CP} & \textbf{SS} & \textbf{LC} & \textbf{EX} & \textbf{VES} \\
         \midrule
         DAC     & 85 & 80 & 79 & 71.00& 67.59 \\
         \textbf{\sysname} & 93 & 83 & 86 & 75.00 & 73.61 \\
    \bottomrule
    \end{tabular}
\end{table}

\begin{figure}[t]
  \begin{tikzpicture}
  \node (table) [inner sep=0pt] {
  \begin{adjustbox}{width=0.98\linewidth}
    \begin{tabular}{l} 
     \\ [-2.0ex]
     \rowcolor{gray!10}\makecell[l]{
     \textbf{Question:} \\
     \textit{What are the first names of the students who live in Haiti}\\
     \textit{permanently or have the cell phone number 09700166582?} \\
     \\ [-1.0ex]
     \textbf{Ground-Truth SQL:} \\
    \textcolor{purple}{SELECT} t1.first\_name \textcolor{purple}{FROM} students AS t1 \textcolor{purple}{JOIN} addresses AS t2 \\
    \textcolor{purple}{ON} t1.permanent\_address\_id = t2.address\_id \\
    \textcolor{purple}{WHERE} t2.country = `Haiti' 
    \textcolor{purple}{OR} t1.cell\_mobile\_number = `09700166582'\\
    \\ [-1.0ex]
    \textbf{CoTE-SQL (LLaMA-3.1-8B):} \\
    \textcolor{purple}{SELECT} students.first\_name \\
    \textcolor{purple} {FROM} students \textcolor{purple}{WHERE} students.permanent\_address\_id IN \\
    ( \textcolor{purple}{SELECT} address\_id \textcolor{purple}{FROM} addresses  \textcolor{purple}{WHERE} addresses.country = `haiti' ) \\
    \textcolor{purple}{OR} students.cell\_mobile\_number = `09700166582'\\
    \textbf{DAC (LLaMA-3.1-8B):} \\
    \textcolor{purple}{SELECT} s.first\_name \textcolor{purple}{FROM} Students s \\
    \textcolor{purple}{WHERE} s.permanent\_address\_id = 5 \\
    \textcolor{purple}{OR} s.cell\_mobile\_number = '09700166582' \\
    }
  \end{tabular}
  \end{adjustbox}
  };

  \end{tikzpicture}
  \caption{Case study for the human evaluation.\label{fig:case_human_eval}}
  \Description{}
  
\end{figure}

\section{Related Work}

\subsection{Prompt-based Reasoning for Text-to-SQL}


Recent advances in text-to-SQL have been driven by LLMs, enabling executable SQL generation from natural language~\cite{zhai2025excot,shen2024select}. CoT prompting~\cite{NEURIPS-CoT} effectively enhances LLM reasoning by decomposing complex queries into intermediate steps using few-shot exemplars. Recent works explore structured prompting to improve SQL generation: DIN-SQL~\cite{NEURIPS2023_72223cc6} decomposes queries via contextual guidance; ACT-SQL~\cite{zhang-etal-2023-act} automates CoT example generation, reducing annotation costs; 
Tai et al.~\cite{tai-etal-2023-exploring} study CoT styles and propose decomposition-based prompting to mitigate error propagation; Xie et al.~\cite{xie-etal-2024-decomposition} divide SQL generation into specialized reasoning modules, advancing CoT performance. 
However, these methods encode reasoning in fixed examples or templates, limiting generalization to complex, compositional queries.

\subsection{Fine-tuning Strategies for Text-to-SQL}

Fine-tuning LLMs on supervised text-to-SQL datasets complements prompting-based methods~\cite{DBLP:conf/iclr/HuSWALWWC22} by training on input-query pairs, improving performance on domain-specific or conversational tasks~\cite{sarker2024enhancing,shen2024improving,wang-etal-2025-mac}. 
MIGA~\cite{Fu_Ou_Yu_Lin_2023} combines multi-task pre-training with SQL perturbations to enhance robustness. FinSQL~\cite{DBLP:conf/sigmod/ZhangMFMG0LL24} uses ChatGPT to synthesize domain-specific datasets for domain-adaptive fine-tuning. 
Hong et al.~\cite{hong-etal-2024-knowledge} inject schema-level knowledge, while Yang et al.~\cite{yang-etal-2024-synthesizing} leverage synthetic and erroneous SQL examples to narrow the gap between open and proprietary LLMs.
Despite their success, most fine-tuning approaches neglect CoT-style reasoning due to limited step-by-step annotations, underperforming on complex or multi-hop SQL queries. Rossiello et al.~\cite{rossiello2025rationalization} distill CoT annotations from a 70B to an 8B model. STaR-SQL~\cite{he-etal-2025-star} combines few-shot prompting with a verifier, with gains mainly from the verifier. Reasoning-SQL~\cite{pourreza2025reasoning} applies partial rewards, and CSC-SQL~\cite{sheng2025csc} separates generation and correction stages optimized via GRPO; however, RL-based methods remain costly, complex, and unstable.

\subsection{Other Explorations in Text-to-SQL}

Several studies aim to reduce cost, latency, and accessibility barriers of LLM-based solutions via architectural and training innovations~\cite{wang2024dac,xie-etal-2024-decomposition,NEURIPS-CoT,DBLP:journals/pvldb/GaoWLSQDZ24,chen2025track}. 
MSC-SQL~\cite{gorti2024msc} selects optimal SQL predictions from small LLMs using multi-sample scoring. 
DTS-SQL~\cite{pourreza-rafiei-2024-dts} applies two-stage fine-tuning for schema linking and SQL generation. Karki et al.~\cite{karki2025smaller} show that 0.5B–3.8B models with parameter-efficient fine-tuning can match 7B models. 
Fan et al.~\cite{fan2024combining} delegate entity linking and sketch generation to small LLMs, using large LLMs for completion. 
SPS-SQL~\cite{yan2025sps} pre-synthesizes schema-aware SQL sketches to improve accuracy. Oliveira et al.~\cite{oliveira2024small} enhance small LLMs with retrieval-augmented generation (RAG).
While these methods improve surface-level generation or alignment, few tackle reasoning limitations. We propose a CoT-enhanced framework for small-scale models, bridging the performance gap through modularized reasoning while maintaining generalization.

\section{Conclusion}
In this paper, we presented \sysname, a framework for enhancing reasoning, generalization, and correctness in LLM-based text-to-SQL generation. By leveraging self-enhanced reasoning traces, structured CoT prompting, and error-aware revision, \sysname addresses key limitations in existing approaches without relying on expensive human annotations. Extensive evaluations demonstrate that our method achieves state-of-the-art performance, particularly excelling on complex and high-difficulty queries. Looking forward, we believe the ideas explored in \sysname offer a broader paradigm for combining self-enhancement, modular reasoning, and feedback-driven correction in LLMs, opening up promising directions for a wide range of complex, multi-step decision-making tasks in natural language interfaces.


\bibliographystyle{ACM-Reference-Format}
\bibliography{ref}

@ARTICLE{10530359,
  author={Zhang, Weixu and Wang, Yifei and Song, Yuanfeng and Wei, Victor Junqiu and Tian, Yuxing and Qi, Yiyan and Chan, Jonathan H. and Wong, Raymond Chi-Wing and Yang, Haiqin},
  journal={IEEE Transactions on Knowledge and Data Engineering}, 
  title={Natural Language Interfaces for Tabular Data Querying and Visualization: A Survey}, 
  year={2024},
  volume={36},
  number={11},
  pages={6699-6718},
  }

@inproceedings{yu-etal-2018-spider,
    title = "{S}pider: A Large-Scale Human-Labeled Dataset for Complex and Cross-Domain Semantic Parsing and Text-to-{SQL} Task",
    author = "Yu, Tao  and
      Zhang, Rui  and
      Yang, Kai  and
      Yasunaga, Michihiro  and
      Wang, Dongxu  and
      Li, Zifan  and
      Ma, James  and
      Li, Irene  and
      Yao, Qingning  and
      Roman, Shanelle  and
      Zhang, Zilin  and
      Radev, Dragomir",
    booktitle = "Proceedings of the 2018 Conference on Empirical Methods in Natural Language Processing",
    year = "2018",
    pages = "3911--3921",
}

@inproceedings{tai-etal-2023-exploring,
    title = "Exploring Chain of Thought Style Prompting for Text-to-{SQL}",
    author = "Tai, Chang-Yu  and
      Chen, Ziru  and
      Zhang, Tianshu  and
      Deng, Xiang  and
      Sun, Huan",
    booktitle = "Proceedings of the 2023 Conference on Empirical Methods in Natural Language Processing",
    year = "2023",
    pages = "5376--5393",
}

@inproceedings{zhang-etal-2023-act,
    title = "{ACT}-{SQL}: In-Context Learning for Text-to-{SQL} with Automatically-Generated Chain-of-Thought",
    author = "Zhang, Hanchong  and
      Cao, Ruisheng  and
      Chen, Lu  and
      Xu, Hongshen  and
      Yu, Kai",
    booktitle = "Findings of the Association for Computational Linguistics: EMNLP 2023",
    year = "2023",
    pages = "3501--3532",
}

@inproceedings{pourreza-rafiei-2024-dts,
    title = "{DTS}-{SQL}: Decomposed Text-to-{SQL} with Small Large Language Models",
    author = "Pourreza, Mohammadreza  and
      Rafiei, Davood",
    booktitle = "Findings of the Association for Computational Linguistics: EMNLP 2024",
    year = "2024",
    pages = "8212--8220",
}

@inproceedings{NEURIPS2023_83fc8fab,
 author = {Li, Jinyang and Hui, Binyuan and Qu, Ge and Yang, Jiaxi and Li, Binhua and Li, Bowen and Wang, Bailin and Qin, Bowen and Geng, Ruiying and Huo, Nan and Zhou, Xuanhe and Chenhao, Ma and Li, Guoliang and Chang, Kevin and Huang, Fei and Cheng, Reynold and Li, Yongbin},
 booktitle = {Advances in Neural Information Processing Systems},
 pages = {42330--42357},
 title = {Can {LLM} Already Serve as A Database Interface? A Big Bench for Large-Scale Database Grounded Text-to-{SQLs}},
 volume = {36},
 year = {2023}
}

@inproceedings{NEURIPS2023_72223cc6,
 author = {Pourreza, Mohammadreza and Rafiei, Davood},
 booktitle = {Advances in Neural Information Processing Systems},
 pages = {36339--36348},
 title = {DIN-SQL: Decomposed In-Context Learning of Text-to-SQL with Self-Correction},
 volume = {36},
 year = {2023}
}

@inproceedings{NEURIPS-CoT,
 author = {Wei, Jason and Wang, Xuezhi and Schuurmans, Dale and Bosma, Maarten and ichter, brian and Xia, Fei and Chi, Ed and Le, Quoc V and Zhou, Denny},
 booktitle = {Advances in Neural Information Processing Systems},
 pages = {24824--24837},
 title = {Chain-of-Thought Prompting Elicits Reasoning in Large Language Models},
 volume = {35},
 year = {2022}
}

@inproceedings{wang-etal-2025-mac,
    title = "{MAC}-{SQL}: A Multi-Agent Collaborative Framework for Text-to-{SQL}",
    author = "Wang, Bing  and
      Ren, Changyu  and
      Yang, Jian  and
      Liang, Xinnian  and
      Bai, Jiaqi  and
      Chai, LinZheng  and
      Yan, Zhao  and
      Zhang, Qian-Wen  and
      Yin, Di  and
      Sun, Xing  and
      Li, Zhoujun",
    booktitle = "Proceedings of the 31st International Conference on Computational Linguistics",
    year = "2025",
    pages = "540--557",
}

@inproceedings{xie-etal-2024-decomposition,
    title = "Decomposition for Enhancing Attention: Improving {LLM}-based Text-to-{SQL} through Workflow Paradigm",
    author = "Xie, Yuanzhen  and
      Jin, Xinzhou  and
      Xie, Tao  and
      Matrixmxlin, Matrixmxlin  and
      Chen, Liang  and
      Yu, Chenyun  and
      Lei, Cheng  and
      Zhuo, Chengxiang  and
      Hu, Bo  and
      Li, Zang",
    booktitle = "Findings of the Association for Computational Linguistics: ACL 2024",
    year = "2024",
    pages = "10796--10816",
}

@inproceedings{hong-etal-2024-knowledge,
    title = "Knowledge-to-{SQL}: Enhancing {SQL} Generation with Data Expert {LLM}",
    author = "Hong, Zijin  and
      Yuan, Zheng  and
      Chen, Hao  and
      Zhang, Qinggang  and
      Huang, Feiran  and
      Huang, Xiao",
    booktitle = "Findings of the Association for Computational Linguistics: ACL 2024",
    year = "2024",
    pages = "10997--11008",
   
}

@inproceedings{yang-etal-2024-synthesizing,
    title = "Synthesizing Text-to-{SQL} Data from Weak and Strong {LLM}s",
    author = "Yang, Jiaxi  and
      Hui, Binyuan  and
      Yang, Min  and
      Yang, Jian  and
      Lin, Junyang  and
      Zhou, Chang",
    booktitle = "Proceedings of the 62nd Annual Meeting of the Association for Computational Linguistics (Volume 1: Long Papers)",
    year = "2024",
    pages = "7864--7875",
}

@article{
Fu_Ou_Yu_Lin_2023, 
title={MIGA: A Unified Multi-Task Generation Framework for Conversational Text-to-SQL}, 
volume={37},   
number={11}, 
journal={Proceedings of the AAAI Conference on Artificial Intelligence},
author={Fu, Yingwen and Ou, Wenjie and Yu, Zhou and Lin, Yue}, 
year={2023},
pages={12790-12798},
}

@inproceedings{DBLP:conf/sigmod/ZhangMFMG0LL24,
  author       = {Chao Zhang and
                  Yuren Mao and
                  Yijiang Fan and
                  Yu Mi and
                  Yunjun Gao and
                  Lu Chen and
                  Dongfang Lou and
                  Jinshu Lin},
  title        = {{FinSQL}: Model-Agnostic LLMs-based Text-to-SQL Framework for Financial
                  Analysis},
  booktitle    = {Companion of the 2024 International Conference on Management of Data,
                  {SIGMOD/PODS} 2024, Santiago, Chile, June 9-15, 2024},
  pages        = {93--105},
  year         = {2024},
}

@article{dubey2024llama,
  title={The llama 3 herd of models},
  author={Dubey, Abhimanyu and Jauhri, Abhinav and Pandey, Abhinav and Kadian, Abhishek and Al-Dahle, Ahmad and Letman, Aiesha and Mathur, Akhil and Schelten, Alan and Yang, Amy and Fan, Angela and others},
  journal={arXiv preprint arXiv:2407.21783},
  year={2024}
}

@article{wang2024dac,
  title={{DAC}: Decomposed Automation Correction for Text-to-{SQL}},
  author={Wang, Dingzirui and Dou, Longxu and Zhang, Xuanliang and Zhu, Qingfu and Che, Wanxiang},
  journal={arXiv preprint arXiv:2408.08779},
  year={2024}
}

@inproceedings{liu-etal-2021-awakening,
    title = "Awakening Latent Grounding from Pretrained Language Models for Semantic Parsing",
    author = "Liu, Qian  and
      Yang, Dejian  and
      Zhang, Jiahui  and
      Guo, Jiaqi  and
      Zhou, Bin  and
      Lou, Jian-Guang",
    booktitle = "Findings of the Association for Computational Linguistics: ACL-IJCNLP 2021",
    year = "2021",
    pages = "1174--1189"
}

@article{DBLP:journals/pvldb/GaoWLSQDZ24,
  author       = {Dawei Gao and
                  Haibin Wang and
                  Yaliang Li and
                  Xiuyu Sun and
                  Yichen Qian and
                  Bolin Ding and
                  Jingren Zhou},
  title        = {Text-to-{SQL} Empowered by Large Language Models: {A} Benchmark Evaluation},
  journal={Proceedings of the VLDB Endowment},
  volume       = {17},
  number       = {5},
  pages        = {1132--1145},
  year         = {2024},
}

@article{hong2024next,
  title={Next-Generation Database Interfaces: A Survey of {LLM}-based {Text-to-SQL}},
  author={Hong, Zijin and Yuan, Zheng and Zhang, Qinggang and Chen, Hao and Dong, Junnan and Huang, Feiran and Huang, Xiao},
  journal={arXiv preprint arXiv:2406.08426},
  year={2024}
}

@article{shen2024select,
  title={SelECT-{SQL}: Self-correcting ensemble Chain-of-Thought for Text-to-{SQL}},
  author={Shen, Ke and Kejriwal, Mayank},
  journal={arXiv preprint arXiv:2409.10007},
  year={2024}
}

@article{yang2024qwen2,
  title={Qwen2.5 technical report},
  author={Yang, An and Yang, Baosong and Zhang, Beichen and Hui, Binyuan and Zheng, Bo and Yu, Bowen and Li, Chengyuan and Liu, Dayiheng and Huang, Fei and Wei, Haoran and others},
  journal={arXiv preprint arXiv:2412.15115},
  year={2024}
}

@article{INR-019,
year = {2009},
volume = {3},
journal = {Found. Trends Inf. Retr.},
title = {The Probabilistic Relevance Framework: BM25 and Beyond},
number = {4},
pages = {333-389},
author = {Stephen Robertson and Hugo Zaragoza}
}

@inproceedings{DBLP:conf/iclr/HuSWALWWC22,
  author       = {Edward J. Hu and
                  Yelong Shen and
                  Phillip Wallis and
                  Zeyuan Allen{-}Zhu and
                  Yuanzhi Li and
                  Shean Wang and
                  Lu Wang and
                  Weizhu Chen},
  title        = {{L}o{RA}: Low-Rank Adaptation of Large Language Models},
  booktitle    = {The Twelfth International Conference on Learning Representations,
                  {ICLR} 2024, Vienna, Austria, May 7-11, 2024},
  year         = {2022},
}

@article{shi2024survey,
  title={A survey on employing large language models for text-to-{SQL} tasks},
  author={Shi, Liang and Tang, Zhengju and Zhang, Nan and Zhang, Xiaotong and Yang, Zhi},
  journal={ACM Computing Surveys},
  year={2024},
}

@inproceedings{ren2024purple,
  title={Purple: Making a large language model a better {SQL} writer},
  author={Ren, Tonghui and Fan, Yuankai and He, Zhenying and Huang, Ren and Dai, Jiaqi and Huang, Can and Jing, Yinan and Zhang, Kai and Yang, Yifan and Wang, X Sean},
  booktitle={2024 IEEE 40th International Conference on Data Engineering (ICDE)}, 
  pages={15--28},
  year={2024},
}

@inproceedings{zhang2025clear,
  title={{CLEAR}: A Parser-Independent Disambiguation Framework for {NL2SQL}},
  author={Zhang, Meng and Ma, Kexin and Xu, Liyang and Zhang, Kedi and Peng, Yuanxi and Jin, Ruochun},
  booktitle={2025 IEEE 41st International Conference on Data Engineering (ICDE)}, 
  pages={3302--3315},
  year={2025},
}

@inproceedings{li2025aid,
  title={{AID-SQL}: Adaptive In-Context Learning of Text-to-{SQL} with Difficulty-Aware Instruction and Retrieval-Augmented Generation},
  author={Li, Xiuwen and Cai, Qifeng and Shu, Yang and Guo, Chenjuan and Yang, Bin},
  booktitle={2025 IEEE 41st International Conference on Data Engineering (ICDE)}, 
  pages={3945--3957},
  year={2025},
}

@inproceedings{fan2023gar,
  title={Gar: A generate-and-rank approach for natural language to {SQL} translation},
  author={Fan, Yuankai and He, Zhenying and Ren, Tonghui and Guo, Dianjun and Chen, Lin and Zhu, Ruisi and Chen, Guanduo and Jing, Yinan and Zhang, Kai and Wang, X Sean},
  booktitle={2023 IEEE 39th International Conference on Data Engineering (ICDE)}, 
  pages={110--122},
  year={2023},
}

@inproceedings{fan2024metasql,
  title={Metasql: A generate-then-rank framework for natural language to {SQL} translation},
  author={Fan, Yuankai and He, Zhenying and Ren, Tonghui and Huang, Can and Jing, Yinan and Zhang, Kai and Wang, X Sean},
  booktitle={2024 IEEE 40th International Conference on Data Engineering (ICDE)}, 
  pages={1765--1778},
  year={2024},
}

@article{liu2024survey,
  title={A Survey of {NL2SQL} with Large Language Models: Where are we, and where are we going?},
  author={Liu, Xinyu and Shen, Shuyu and Li, Boyan and Ma, Peixian and Jiang, Runzhi and Zhang, Yuxin and Fan, Ju and Li, Guoliang and Tang, Nan and Luo, Yuyu},
  journal={arXiv preprint arXiv:2408.05109},
  year={2024}
}

@article{fan2024grounding,
  title={Grounding Natural Language to {SQL} Translation with Data-Based Self-Explanations},
  author={Fan, Yuankai and Ren, Tonghui and Huang, Can and He, Zhenying and Wang, X Sean},
  journal={arXiv preprint arXiv:2411.02948},
  year={2024}
}

@article{liu2025nl2sql,
  title={Nl2sql-bugs: A benchmark for detecting semantic errors in {NL2SQL} translation},
  author={Liu, Xinyu and Shen, Shuyu and Li, Boyan and Tang, Nan and Luo, Yuyu},
  journal={arXiv preprint arXiv:2503.11984},
  year={2025}
}

@inproceedings{chen2025track,
  title={{Track-SQL}: Enhancing Generative Language Models with Dual-Extractive Modules for Schema and Context Tracking in Multi-turn Text-to-{SQL}},
  author={Chen, Bingfeng and Shi, Shaobin and Luo, Yongqi and Xu, Boyan and Cai, Ruichu and Hao, Zhifeng},
  booktitle={Proceedings of the 2025 Conference of the Nations of the Americas Chapter of the Association for Computational Linguistics: Human Language Technologies (Volume 1: Long Papers)},

  pages={10690--10708},
  year={2025}
}

@article{zhao2024sphinteract,
  title={Sphinteract: Resolving Ambiguities in {NL2SQL} through User Interaction},
  author={Zhao, Fuheng and Deep, Shaleen and Psallidas, Fotis and Floratou, Avrilia and Agrawal, Divyakant and Abbadi, Amr El},
  journal={Proceedings of the VLDB Endowment},
  volume={18},
  number={4},
  pages={1145--1158},
  year={2024},
}

@article{sarker2024enhancing,
  title={Enhancing {LLM} Fine-tuning for Text-to-{SQLs} by {SQL} Quality Measurement},
  author={Sarker, Shouvon and Dong, Xishuang and Li, Xiangfang and Qian, Lijun},
  journal={arXiv preprint arXiv:2410.01869},
  year={2024}
}

@article{zhai2025excot,
  title={{ExCoT}: Optimizing reasoning for text-to-{SQL} with execution feedback},
  author={Zhai, Bohan and Xu, Canwen and He, Yuxiong and Yao, Zhewei},
  journal={arXiv preprint arXiv:2503.19988},
  year={2025}
}

@article{shen2024improving,
  title={Improving Retrieval-augmented Text-to-{SQL} with {AST}-based Ranking and Schema Pruning},
  author={Shen, Zhili and Vougiouklis, Pavlos and Diao, Chenxin and Vyas, Kaustubh and Ji, Yuanyi and Pan, Jeff Z},
  journal={arXiv preprint arXiv:2407.03227},
  year={2024}
}

@inproceedings{huang2024data,
  title={Data-Centric Text-to-{SQL} with Large Language Models},
  author={Huang, Zezhou and Zhang, Shuo and Liu, Kechen and Wu, Eugene},
  booktitle={NeurIPS 2024 Third Table Representation Learning Workshop},
  year={2024},
}

@article{luo2024ptd,
  title={{PTD-SQL}: Partitioning and targeted drilling with {LLMs} in Text-to-{SQL}},
  author={Luo, Ruilin and Wang, Liyuan and Lin, Binghuai and Lin, Zicheng and Yang, Yujiu},
  journal={arXiv preprint arXiv:2409.14082},
  year={2024}
}

@article{gorti2024msc,
  title={{MSC-SQL}: Multi-sample critiquing small language models for text-to-{SQL} translation},
  author={Gorti, Satya Krishna and Gofman, Ilan and Liu, Zhaoyan and Wu, Jiapeng and Vouitsis, No{\~A}{\c{G}}l and Yu, Guangwei and Cresswell, Jesse C and Hosseinzadeh, Rasa},
  journal={arXiv preprint arXiv:2410.12916},
  year={2024}
}

@inproceedings{karki2025smaller,
  title={Smaller Large Language Models for Text-to-{SQL}: Performance Analysis and Optimal Performance},
  author={Karki, Sujan and Karki, Pukar and Shrestha, Binay Lal and Jha, Tantra Nath},
  booktitle={2025 International Conference on Inventive Computation Technologies (ICICT)},
  pages={1--7},
  year={2025},
}

@article{fan2024combining,
  title={Combining small language models and large language models for zero-shot {NL2SQL}},
  author={Fan, Ju and Gu, Zihui and Zhang, Songyue and Zhang, Yuxin and Chen, Zui and Cao, Lei and Li, Guoliang and Madden, Samuel and Du, Xiaoyong and Tang, Nan},
  journal={Proceedings of the VLDB Endowment},
  volume={17},
  number={11},
  pages={2750--2763},
  year={2024},
}

@article{yan2025sps,
  title={{SPS-SQL}: Enhancing Text-to-{SQL} generation on small-scale {LLMs} with pre-synthesized queries},
  author={Yan, Liang and Wan, Qichen and Liu, Chuanyi and Duan, Shaoming and Han, Peiyi and Xu, Yong},
  journal={Pattern Recognition Letters},
  year={2025},
}

@inproceedings{oliveira2024small,
  title={Small, Medium, and Large Language Models for Text-to-{SQL}},
  author={Oliveira, Aiko and Nascimento, Eduardo and Pinheiro, Jo{\~a}o and Avila, Caio Viktor S and Coelho, Gustavo and Feij{\'o}, Lucas and Izquierdo, Yenier and Garc{\'\i}a, Grettel and Leme, Luiz Andr{\'e} P Paes and Lemos, Melissa and others},
  booktitle={International Conference on Conceptual Modeling},
  pages={276--294},
  year={2024},
}

@article{guo2025deepseek,
  title={Deepseek-r1: Incentivizing reasoning capability in llms via reinforcement learning},
  author={Guo, Daya and Yang, Dejian and Zhang, Haowei and Song, Junxiao and Zhang, Ruoyu and Xu, Runxin and Zhu, Qihao and Ma, Shirong and Wang, Peiyi and Bi, Xiao and others},
  journal={arXiv preprint arXiv:2501.12948},
  year={2025}
}

@article{pourreza2025reasoning,
  title={Reasoning-sql: Reinforcement learning with sql tailored partial rewards for reasoning-enhanced text-to-sql},
  author={Pourreza, Mohammadreza and Talaei, Shayan and Sun, Ruoxi and Wan, Xingchen and Li, Hailong and Mirhoseini, Azalia and Saberi, Amin and Arik, Sercan and others},
  journal={arXiv preprint arXiv:2503.23157},
  year={2025}
}

@article{sheng2025csc,
  title={CSC-SQL: Corrective Self-Consistency in Text-to-SQL via Reinforcement Learning},
  author={Sheng, Lei and Xu, Shuai-Shuai},
  journal={arXiv preprint arXiv:2505.13271},
  year={2025}
}

@inproceedings{he-etal-2025-star,
    title = "{ST}a{R}-{SQL}: Self-Taught Reasoner for Text-to-{SQL}",
    author = "He, Mingqian  and
      Shen, Yongliang  and
      Zhang, Wenqi  and
      Peng, Qiuying  and
      Wang, Jun  and
      Lu, Weiming",
    booktitle = "Proceedings of the 63rd Annual Meeting of the Association for Computational Linguistics (Volume 1: Long Papers)",
    month = jul,
    year = "2025",
    pages = "24365--24375",
}

@article{rossiello2025rationalization,
  title={Rationalization Models for Text-to-SQL},
  author={Rossiello, Gaetano and Pham, Nhan and Glass, Michael and Lee, Junkyu and Subramanian, Dharmashankar},
  journal={arXiv preprint arXiv:2502.06759},
  year={2025}
}

@inproceedings{zheng2023rhb,
  title={RHB-Net: A Relation-aware Historical Bridging Network for Text2SQL Auto-Completion},
  author={Zheng, Bolong and Bi, Lei and Xi, Ruijie and Chen, Lu and Gao, Yunjun and Zhou, Xiaofang and Jensen, Christian S},
  booktitle={Proceedings of the 46th International ACM SIGIR Conference on Research and Development in Information Retrieval},
  pages={1458--1467},
  year={2023}
}

@inproceedings{gong2024graph,
  title={Graph reasoning enhanced language models for text-to-sql},
  author={Gong, Zheng and Sun, Ying},
  booktitle={Proceedings of the 47th International ACM SIGIR Conference on Research and Development in Information Retrieval},
  pages={2447--2451},
  year={2024}
}

@article{yang2026iquest,
  title={IQuest-Coder-V1 Technical Report},
  author={Yang, Jian and Zhang, Wei and Guo, Shawn and Ye, Zhengmao and Jing, Lin and Liu, Shark and Li, Yizhi and Wu, Jiajun and Liu, Cening and Ma, X and others},
  journal={arXiv preprint arXiv:2603.16733},
  year={2026}
}

\end{document}